\documentclass{article} %
\PassOptionsToPackage{table}{xcolor}
\usepackage{iclr2026_conference,times}

\usepackage{hyperref}
\usepackage{url}
\usepackage{amsmath}
\usepackage{amssymb}
\usepackage{mathtools}
\usepackage{amsthm}
\usepackage{subcaption}
\usepackage{caption}

\usepackage{algorithm}
\usepackage{algorithmic}
\usepackage{booktabs}
\usepackage{multirow}
\usepackage{enumitem}
\usepackage{graphicx}
\usepackage{wrapfig}
\usepackage{amsfonts}       %
\usepackage{nicefrac}       %
\usepackage{microtype}      %
\usepackage{xcolor}
\definecolor{motifgreen}{HTML}{88BDA4}

\title{Fraglingo: Open-Vocabulary Molecular Design via Attachment-Aware Fragment Generation}

\author{Thao Nguyen \& Jeonghwan Kim \& Zhenhailong Wang \& Heng Ji \\
Siebel School of Computing and Data Science \\
University of Illinois Urbana-Champaign \\
Urbana, IL 61801, USA \\
\texttt{\{thaotn2,jk100,wangz3,hengji\}@illinois.edu}
}

\iclrfinalcopy %
\begin{document}

\maketitle

\begin{abstract}
We introduce \textbf{Fraglingo}, an autoregressive molecular generator that constructs molecules step by step from \emph{fragments} --- chemically meaningful substructures connected through predefined attachment sites. At each generation step, the current partial molecule is the \emph{growing molecule}, and one of its available attachment sites is selected as the \emph{active attachment site} from which generation continues. Fraglingo jointly predicts what fragment to add and how it should attach by predicting an \emph{attachment-aware fragment embedding}, a continuous representation that encodes both fragment identity and attachment configuration, and retrieving the nearest fragment through latent-space search. To capture attachment context, we introduce a \emph{wildcard-anchored readout} that represents both the growing molecule and candidate fragments relative to their attachment sites, enabling a single latent prediction to determine both which fragment to attach and where the new bond should form. Because prediction operates in a continuous embedding space rather than over fixed fragment identifiers, larger fragment libraries can be introduced at inference time without retraining. This retrieval-based formulation provides a unified generation primitive for molecule generation, scaffold generation, scaffold decoration, and molecule optimization. Fraglingo also supports \emph{property-conditional generation}, where desired molecular properties are provided as generation conditions to control the properties of generated molecules. On controlled property-conditional benchmarks, Fraglingo achieves stronger joint property control than comparably trained baselines while maintaining competitive validity, uniqueness, and novelty, and generalizes to fragment libraries up to $4\times$ larger than those used during training without retraining. \textit{Code is available at:} \url{https://anonymous.4open.science/r/FragLingo-3551}.
\end{abstract}

\section{Introduction}

Molecular design aims to discover chemical structures with desired biological and physicochemical properties, and is a central problem in drug discovery. Generative models provide a way to search this enormous chemical space by learning how to construct new molecules while satisfying objectives such as potency, drug-likeness, or similarity to a known scaffold. A fundamental design choice is therefore the \emph{unit of generation}: what does the model predict at each step? Existing approaches generate molecules as sequences of SMILES~\citep{weininger1988smiles} or SELFIES~\citep{krenn2020self} tokens, atoms and bonds in molecular graphs, functional modules~\citep{mCLM2026}, or entire molecular graphs through diffusion and flow models~\citep{segler2018generating,gomez2018automatic,ross2022large,shi2020graphaf,hoogeboom2022equivariant}. These representations are expressive, but their generation units are often lower-level than the structural changes considered during molecular design. For example, changing or misplacing a single parenthesis or ring-closure token in SMILES can alter the molecular connectivity or make the string invalid, while atom-by-atom graph generation may require many sequential decisions to construct a single recognizable chemical motif.

Fragment-based generation instead constructs molecules from larger, chemically meaningful substructures. This more closely resembles common medicinal-chemistry operations such as adding, replacing, extending, or decorating molecular motifs~\citep{hajduk2007decade,kirsch2019concepts}. A molecule is first decomposed into \emph{fragments}, with the broken bonds represented as \emph{attachment sites}. Generation then proceeds autoregressively: given a partially constructed \emph{growing molecule}, the model selects an open attachment site and predicts a fragment to connect there. Table~\ref{tab:terminology} summarizes the terminology used throughout this work.

\
\begin{table}[t]
\centering
\small
\caption{Key terminology used in Fraglingo.}
\vspace{-0.5em}
\label{tab:terminology}
\begin{tabular}{p{0.25\linewidth}p{0.68\linewidth}}
\toprule
\textbf{Term} & \textbf{Definition} \\
\midrule
Fragment & A chemically meaningful molecular substructure used as a generation unit. \\
Attachment site & A position on a fragment where a bond to another fragment can be formed, represented by a wildcard atom (*). \\
Active attachment site & The open attachment site of the growing molecule from which the next generation step proceeds. \\
Growing molecule & The partially assembled molecule at the current generation step. \\
Attachment-aware & A representation conditioned on the specific attachment site through which a fragment participates in assembly. \\
Wildcard-anchored readout & A graph readout that represents a molecule relative to a selected wildcard attachment site. \\
\bottomrule
\end{tabular}
\vspace{-1.6em}
\end{table}

\begin{figure}[!t]
    \centering
    \includegraphics[width=0.9\linewidth]{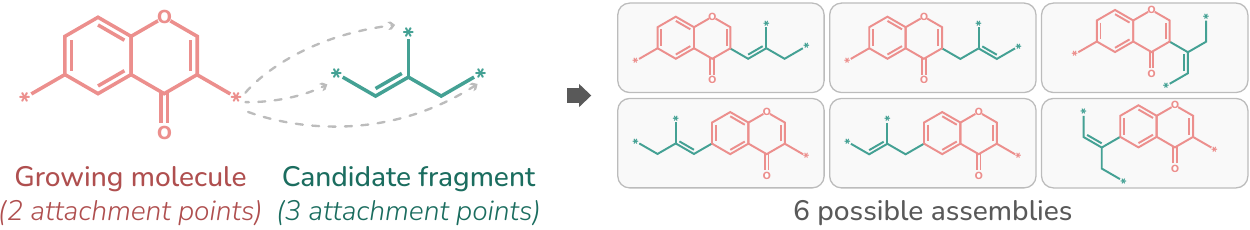}
    \caption{Attachment ambiguity in fragment-based molecular generation. The same pair of fragments can produce different molecular structures depending on which attachment sites are connected.}
    \vspace{-1.2em}
    \label{fig:example}
\end{figure}

\vspace{-1.6em}Despite its advantages, fragment-based generation introduces two fundamental challenges. First, many existing approaches formulate next-fragment prediction as classification over a fixed vocabulary~\citep{jin2020hierarchical,kong2022molecule,yue2024unlocking}. The model can therefore directly select only fragments represented in its training-time output space, making it difficult to exploit larger or updated fragment libraries at inference time. Second, predicting \emph{which} fragment to use is not sufficient: the model must also determine \emph{how} it attaches. A growing molecule may contain several open attachment sites, and a candidate fragment may itself have multiple possible attachment points. As illustrated in Figure~\ref{fig:example}, the same pair of fragments can therefore produce different molecular structures depending on which attachment sites are connected. A global representation of either molecule does not explicitly distinguish these local attachment configurations.

We introduce \textbf{Fraglingo}, an attachment-aware autoregressive fragment generator designed to address both challenges. Fraglingo decomposes molecules into fragments and repeatedly predicts the next fragment to attach to the growing molecule. The framework is agnostic to the fragmentation scheme; we use BRICS~\citep{degen2008art} as the primary scheme and BFE~\citep{nguyen2026mollingo} in an ablation study. Rather than predicting a discrete fragment identifier, Fraglingo performs \emph{continuous latent retrieval}: the model predicts an embedding for the desired next fragment and retrieves its nearest neighbor from a fragment embedding library. The output space is therefore defined by learned chemical representations rather than fixed class identifiers, allowing the retrieval library to be expanded at inference time without retraining.

The key challenge is making this retrieval \emph{attachment-aware}. Fraglingo introduces a \emph{wildcard-anchored readout} that constructs representations relative to a selected attachment site. The growing molecule is encoded from the perspective of its active attachment site, and candidate fragments are represented from the perspective of each possible attachment site. Retrieval therefore asks a local and chemically explicit question: \emph{which fragment, viewed through which attachment site, best continues this particular site of the growing molecule?} A single retrieved representation consequently determines both the fragment identity and its attachment configuration. Desired molecular properties can additionally be supplied as conditions, allowing the same generation process to steer molecules toward specified property profiles.

This attachment-aware retrieval primitive supports four molecular design tasks within the same framework: molecule generation, scaffold generation, scaffold decoration, and molecule optimization. Molecules and scaffolds are constructed through autoregressive fragment assembly, scaffolds are decorated by retrieving branches at specified attachment sites, and existing molecules are optimized through conditioned fragment replacement. Across these settings, Fraglingo retains the same basic operation: predict an attachment-aware representation and retrieve a fragment from a continuous embedding space.

In summary, our contributions are:
\begin{itemize}[leftmargin=*]
    \item \textbf{Wildcard-anchored readout for attachment-aware generation.} We introduce a site-specific graph readout that represents both the growing molecule and candidate fragments relative to their attachment sites, enabling joint prediction of \emph{what} fragment to add and \emph{where} to attach it.
    \item \textbf{Continuous latent retrieval for open-vocabulary fragment generation.} We formulate next-fragment prediction as retrieval in a learned embedding space rather than classification over fixed fragment identifiers, enabling the fragment library to be expanded at inference time without retraining.
    \item \textbf{One retrieval primitive for four molecular design tasks.} Fraglingo supports molecule generation, scaffold generation, scaffold decoration, and molecule optimization within a unified fragment-retrieval framework.
\end{itemize}

\vspace{-0.8em}
\section{Molecular Design via Latent-Space Autoregressive Fragment Generation}
\vspace{-0.8em}

\label{sec:method}

This section first introduces the common generation primitive used by Fraglingo, then explains how the same primitive is adapted to molecule generation, molecule optimization, scaffold generation, and scaffold decoration. Detailed model design, training objectives, and configuration choices are provided in Appendix~\ref{app:model}.

\vspace{-0.8em}

\subsection{Molecule / Scaffold Generation}
\label{sec:fraggen}

\begin{figure}[!b]
    \centering
    \includegraphics[width=0.92\linewidth]{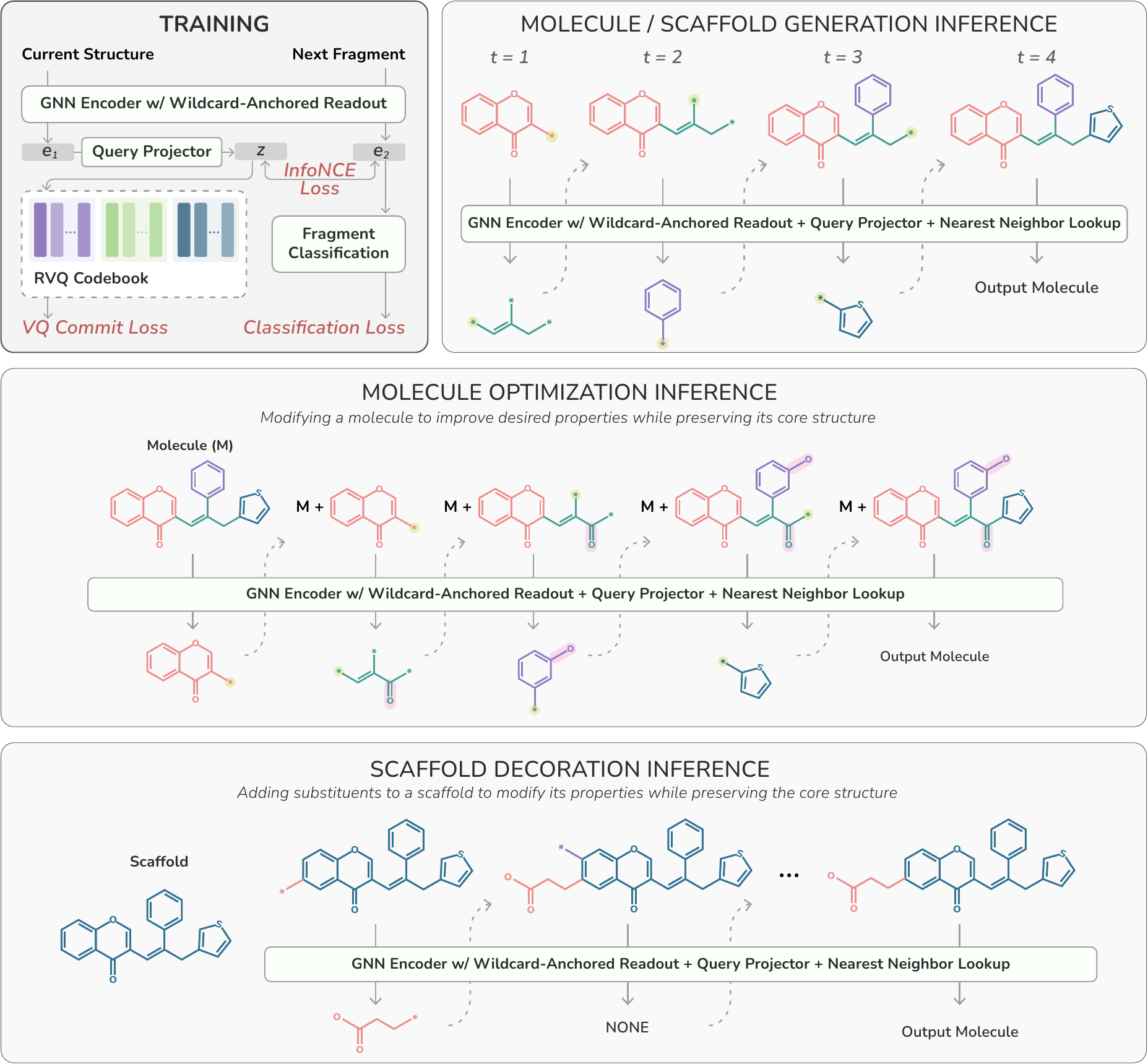}
    \caption{Overview of Fraglingo training and inference across four molecular design tasks. The architecture shown illustrates unconditional generation, with the GNN encoder and query projector abbreviated for clarity; detailed architectures are provided in Figure~\ref{fig:model}.
    }
    \label{fig:teaser}
\end{figure}

\vspace{-0.5em}
\paragraph{Fragment assembly as next-fragment prediction.} Fraglingo generates a molecule by sequentially assembling chemically meaningful fragments. We first decompose each training molecule using BRICS, which breaks selected bonds and replaces them with wildcard attachment atoms. This produces a fragment tree $\{F_1,\ldots,F_T\}$. Starting from a cap fragment, assembly proceeds in breadth-first order. At step $t$, the model observes a partially assembled molecule $G_t$ and an active wildcard $w_t$, and must predict the next fragment $F_{t+1}$ and the wildcard through which it should attach. Scaffold generation follows exactly the same process, with Murcko scaffolds~\citep{bemis1996properties} used as the training structures instead of complete molecules. Figure~\ref{fig:teaser} illustrates this generation process.

\vspace{-0.7em}
\paragraph{Why attachment-aware representations?} A fragment cannot be represented only by its molecular identity because the same fragment may contain multiple attachment sites that lead to different assemblies. Fraglingo therefore represents both the growing molecule and candidate fragments from the perspective of a specific wildcard. A shared graph encoder $\phi$, implemented as a 6-layer FragmentGPS network based on GPS~\citep{rampavsek2022recipe}, first computes atom representations. The designated wildcard then acts as an attention query over the real atoms of the graph, producing the \emph{wildcard-anchored readout}:
$e_1=\phi(G_t,w_t)$ and $e_2=\phi(F_{t+1},w_{t+1})$,
where $e_1$ represents the growing molecule from the active attachment site and $e_2$ represents the candidate fragment from the site through which it would attach. Consequently, a fragment with multiple wildcards has a different representation for each possible attachment site. This allows the prediction to specify both \emph{which fragment} to retrieve and \emph{where} that fragment should attach.

\vspace{-0.7em}
\paragraph{Why continuous fragment retrieval?} Treating next-fragment prediction as classification restricts inference to a fixed set of fragment classes. Fraglingo instead predicts a point in the learned fragment embedding space. Desired molecular properties $\mathbf{p}$ are first incorporated into the attachment-aware context,
$e_1^{\mathrm{cond}}=e_1+W_p\mathbf{p}+\mathbf{b}_p,$
where we condition on logP, MW, QED, TPSA, HBD, HBA, and RotBonds computed with RDKit~\citep{rdkit}. A two-layer MLP then maps this context to the predicted next-fragment embedding:
$\hat{z}=f(e_1^{\mathrm{cond}}).$
Property conditions are randomly masked with probability $0.2$ during training to enable classifier-free guidance at inference.

\vspace{-0.7em}
\paragraph{Learning the retrieval space.} Training must make $\hat{z}$ close to the correct attachment-aware fragment embedding while maintaining a structured and discriminative fragment space. We jointly optimize
\begin{equation}
\label{eq:main-loss}
\mathcal{L}=\lambda_{\mathrm{NCE}}\mathcal{L}_{\mathrm{InfoNCE}}(\hat{z},e_2)+\lambda_{\mathrm{cls}}\mathcal{L}_{\mathrm{cls}}(e_2)+\lambda_{\mathrm{commit}}\mathcal{L}_{\mathrm{commit}},
\end{equation}
where InfoNCE~\citep{oord2018representation} aligns the prediction with the correct fragment embedding, the classification objective prevents the fragment space from collapsing, and the commitment loss regularizes the predicted embedding using a residual vector quantizer (RVQ). The RVQ contains four stages with 64 codes per stage and is used only during training. Grid search selects $\lambda_{\mathrm{NCE}}:\lambda_{\mathrm{cls}}:\lambda_{\mathrm{commit}}=1:0.5:0.25$.
\vspace{-0.7em}
\paragraph{Open-vocabulary inference.} After training, both the classification head and RVQ are discarded. We encode every candidate fragment with the trained graph encoder and construct an attachment-aware embedding lookup table. At each generation step, Fraglingo predicts $\hat{z}$ and retrieves its nearest fragment embedding using cosine similarity; the corresponding fragment is then connected to the active wildcard. Because fragment selection depends on embeddings produced by the encoder rather than fixed output classes, new fragments can be encoded and added to the retrieval library after training without retraining the generator. Generation repeats until no open attachment sites remain.

\vspace{-0.8em}
\subsection{Molecule Optimization}
\vspace{-0.6em}
\label{sec:optimization}

Molecule optimization asks for local edits that improve target properties while preserving much of the starting molecule. To condition generation on the molecule being optimized, we add a global reference encoder. This encoder summarizes the starting molecule so that the model can choose edits that are compatible with the original structure. Since a complete reference molecule has no open attachment point, we replace the wildcard-anchored readout with a learned-query attention pool (GlobalPool): a trainable query vector attends over all atoms of the reference molecule, yielding a global embedding $e_{\text{ref}}$. The predicted next-fragment embedding becomes:
$\label{eq:reference-cond}
    \hat{z} = f\!\left(e_1 + W_p \mathbf{p} + \mathbf{b}_p + W_r e_{\text{ref}}\right)$,
so generation is jointly conditioned on the current partial assembly, target properties, and the reference structure. During training, $\mathbf{p}$ and $e_{\text{ref}}$ are each independently dropped out: dropping $e_{\text{ref}}$ recovers plain conditional generation, while dropping $\mathbf{p}$ recovers reference-anchored generation without a property target. 
The optimized molecule is generated with the same fragment-retrieval process as molecule generation. Substitutions, insertions, and deletions all emerge naturally through the sequence of fragment predictions. For example, a deletion corresponds to generating a fragment tree with fewer nodes than the reference molecule. Similarity to the original molecule is not imposed by a hand-coded constraint; instead, the model learns the expected degree of preservation from paired training examples through the reference-conditioning signal.

\vspace{-0.8em}
\subsection{Scaffold Decoration}
\vspace{-0.6em}
\label{sec:decoration}

Scaffold decoration uses the same retrieval primitive in a different setting. Given a fixed Murcko scaffold~\citep{bemis1996properties}, the task is to independently predict which branch fragment should attach at each growable site. Unlike autoregressive generation, decoration is a per-site retrieval problem; the scaffold never changes, and each site is filled in a single forward pass.
The only architectural change is the readout query. Instead of treating a wildcard as the active attachment site, the query is centered on the scaffold atom at the growable site. The attachment-anchored readout (Eq.~\ref{eq:readout}) lets this atom attend to the full scaffold context, producing a representation that asks \textit{which branch best fits this attachment site in this scaffold}. Candidate branches are drawn from a vocabulary of \texttt{*}-R fragments, where \texttt{*} denotes the attachment atom, together with a learned \texttt{[NONE]} token that leaves the site undecorated.
Property conditioning and the $\mathrm{MLP}_{\mathrm{Mapper}}$ predictor $f$ are identical to those used for molecule generation. At inference, $\hat{z}=f(e_1^{\mathrm{cond}})$ is predicted independently for each growable site from the fixed, undecorated scaffold, and the branch whose embedding is nearest to $\hat{z}$ is retrieved from the full branch vocabulary by cosine similarity.

\vspace{-0.8em}
\section{Experiments}
\vspace{-0.8em}
\label{sec:experiments}

\subsection{Training Data}
\vspace{-0.6em}
\label{sec:data}

We construct task-specific grids of datasets to study how Fraglingo scales with both vocabulary size and corpus size. Molecules are first decomposed into fragments using the BRICS algorithm~\citep{degen2008art}, and the fragment vocabulary is defined as the $k$ most frequent BRICS fragments in the training corpus. Each grid is formed by choosing a vocabulary size $k$, retaining examples whose constituent fragments are fully covered by that vocabulary, and then subsampling a corpus of size $N$. The resulting $k \times N$ design lets us separate the effect of a larger retrieval library from the effect of more training examples while keeping the data-construction protocol consistent across tasks.
All datasets derive from a shared base corpus of 883M commercially available molecules from ZINC15~\citep{sterling2015zinc}, filtered to molecular weight below 500 Da and retaining only molecules that decompose into at least two BRICS fragments. Task-specific datasets are constructed from this corpus as described below. In all cases, seven molecular properties are computed per molecule using RDKit~\citep{rdkit} for conditioning. Some of these descriptors are approximately additive over fragments, such as MW and hydrogen-bond counts, whereas others, such as logP, QED, TPSA, and RotBonds, depend more strongly on the assembled molecular structure and therefore provide complementary conditioning signals; Appendix~\ref{app:property-descriptors} summarizes each descriptor. Datasets are divided into train, validation, and test scaffold splits with validation and test held out, using an 8:1:1 ratio.

\textit{Molecule Generation.}
The molecule-generation datasets use fragment vocabularies of size $k \in \{100, 500, 1000, 2000\}$ and training corpus sizes $N \in \{1\text{k}, 10\text{k}, 100\text{k}, 1\text{M}\}$.
\textit{Scaffold Generation.}
We extract Murcko scaffolds from the base molecules, yielding approximately 71M unique scaffolds. Following the protocol above, we use scaffold fragment vocabularies of size $k \in \{50, 100, 200, 500\}$ and training corpus sizes $N \in \{10\text{k}, 100\text{k}, 1\text{M}\}$. The smaller vocabulary range reflects the reduced fragment diversity in scaffold space.
\textit{Molecule Optimization.}
We construct matched molecular pairs from the base corpus using Tanimoto similarity in the range $[0.70, 0.95]$, ensuring that each pair shares a common scaffold while differing by a small structural modification. Pairs are further filtered by aligning their BRICS fragment trees and requiring that the transformation corresponds to the insertion, deletion, or substitution of one or two fragments. Following the construction protocol above, we use fragment vocabularies of size $k \in \{500, 1000, 2000\}$ and training corpus sizes $N \in \{10\text{k}, 100\text{k}, 1\text{M}\}$.
\textit{Scaffold Decoration.}
For scaffold decoration, each molecule is decomposed into a Murcko scaffold and the branches attached to that scaffold. We rank extracted branches by frequency and retain the top-$k$ branches as the decoration vocabulary, with $k \in \{500, 1000, 2000\}$. We keep only molecules whose attached branches are all covered by this vocabulary, subsample $N \in \{10\text{k}, 100\text{k}, 1\text{M}\}$ molecules, and decompose the retained molecules into scaffold-plus-branch training examples: the scaffold provides the fixed core for decoration, and the attached branches serve as the ground-truth retrieval targets.

\vspace{-0.8em}
\subsection{Benchmarking Setup}
\vspace{-0.6em}
\label{subsec:eval_cond}

We choose baselines with publicly available implementations that can be trained under a unified experimental protocol. Existing open-source methods primarily target molecule generation and molecule optimization; to the best of our knowledge, no publicly available baselines support scaffold generation or scaffold decoration. Accordingly, we benchmark Fraglingo against prior work on the two tasks where direct comparison is possible, while evaluating scaffold generation and scaffold decoration through comprehensive self-evaluation, including generation quality and controlled studies of vocabulary size and training corpus size.
Every model is trained with three independent random seeds, and all reported results are the mean and standard deviation across seeds. All evaluated models, including Fraglingo, contain between 1.8M and 6.4M trainable parameters.

\vspace{-0.7em}
\paragraph{Property-Conditional Molecule Generation.}

We compare against MolGPT~\citep{bagal2021molgpt}, MolRWKV~\citep{li2025molrwkv}, and PS-VAE~\citep{kong2022molecule}.
The selected baselines span complementary generation paradigms, including SMILES autoregression, graph-enhanced sequence generation, and fragment or subgraph latent-variable modeling.
To ensure a controlled comparison, all trainable models are retrained on the same 10k-molecule training corpus used by Fraglingo. Fragment-based methods use a shared BRICS vocabulary of 500 fragments, while non-fragment methods are trained on the corresponding molecular dataset under their native representation. During evaluation, each model generates 20 molecules for each of 100 randomly sampled target-property vectors, conditioning jointly on seven molecular properties (Appendix~\ref{app:property-descriptors}).

\vspace{-0.7em}
\paragraph{Molecule Optimization.}

We compare against MOLLEO~\citep{wang2025efficient}, HierVAE~\citep{jin2020hierarchical}, and JT-VAE~\citep{jin2018junction}, retraining all methods under the same experimental protocol. All models are trained on the same 10k matched-pair dataset using a BRICS fragment vocabulary of 500 fragments, with an 8:1:1 scaffold split and three random seeds.
Unlike unconditional property generation, optimization targets must remain chemically close to the reference molecule. Randomly sampling target-property vectors frequently produces unattainable objectives or requires large structural changes. Thus, for each held-out optimization pair, we use the original molecule together with the target properties of its paired optimized molecule as input. This guarantees that a chemically plausible solution exists while preserving the intended optimization setting and avoiding unrealistic property shifts.

\vspace{-0.8em}
\subsection{Evaluation Metrics}
\vspace{-0.6em}
\label{subsec:evaluation_metrics}

\paragraph{Generation quality.}
We report four standard sanity-check metrics that assess whether a model produces valid, varied, and non-memorized molecules, independent of property control.
\textit{Validity} measures the fraction of generated outputs that parse as valid RDKit~\citep{rdkit} molecules.
\textit{Uniqueness} is the fraction of valid molecules with distinct SMILES strings. \textit{Novelty} is the fraction of unique molecules absent from the training set.
\textit{Diversity} is defined as $1 - \overline{T}$, where $\overline{T}$ is the mean pairwise Tanimoto similarity over Morgan fingerprints (radius 2, 2048 bits), averaged within each condition batch and then across conditions.

\vspace{-0.7em}
\paragraph{Property control.}
We assess property control at the per-property, joint, and partial levels.

\textit{Single-property conditioning.} Each of the seven molecular properties is evaluated independently by conditioning the model on a single target property. For each property, we measure the Spearman correlation between the requested target value and the value of the generated molecule computed using RDKit descriptors.

\textit{Normalized Joint Distance (NJD).} The primary multi-property metric is the Normalized Joint Distance, defined per molecule as:
\begin{equation}
    \text{NJD} = \sqrt{\sum_{p} \left(\frac{\hat{y}_p - y_p}{\sigma_p}\right)^2}
\end{equation}
where $\hat{y}_p$ and $y_p$ are the achieved and target values for property $p$, and $\sigma_p$ is the training-set standard deviation of property $p$. Normalizing by $\sigma_p$ prevents high-magnitude properties (e.g., MW in daltons) from dominating over unit-bounded ones (e.g., QED $\in [0,1]$). Lower NJD is better.

\textit{Joint success rate.} We report the fraction of generated molecules where all seven properties simultaneously fall within per-property absolute tolerances: logP $\pm 1.0$, MW $\pm 50$ Da, QED $\pm 0.15$, TPSA $\pm 20$, HBD $\pm 1$, HBA $\pm 2$, RotBonds $\pm 2$. We evaluate at two tolerance multipliers ($1\times, 2\times$); Joint@$1\times$ (unscaled) is the stricter number.

\textit{Partial satisfaction.} Because the all-or-nothing joint metric can obscure near-misses, we also report the mean count of properties individually within their $1\times$ threshold (out of 7). This captures generations that are close but fail on a single property.

\vspace{-0.8em}
\subsection{Results}
\vspace{-0.6em}
\label{subsec:results}

Section~\ref{subsec:wildcard-readout-geometry} probes the learned representation and shows that Fraglingo captures an attachment-aware view of the molecular context.
Section~\ref{subsec:conditional-generation-results} shows that Fraglingo achieves the strongest joint property control among comparably trained baselines. Section~\ref{subsec:vocab-expansion} then demonstrates inference-time vocabulary expansion, the capability that continuous retrieval uniquely enables. Appendix~\ref{appendix:component-ablation} isolates the contribution of the continuous-retrieval objective. Appendix~\ref{1:molecule-optimization-results},~\ref{2:scaffold-generation-results}~and~\ref{3:scaffold-decoration-results} show that the same primitive transfers to molecule optimization, scaffold generation, and scaffold decoration without modification.

\vspace{-0.6em}
\subsubsection{The Wildcard Anchor Learns a Attachment-Aware Representation}
\vspace{-0.4em}
\label{subsec:wildcard-readout-geometry}

\begin{figure}[!hbt]
  \centering
  \includegraphics[width=\textwidth]{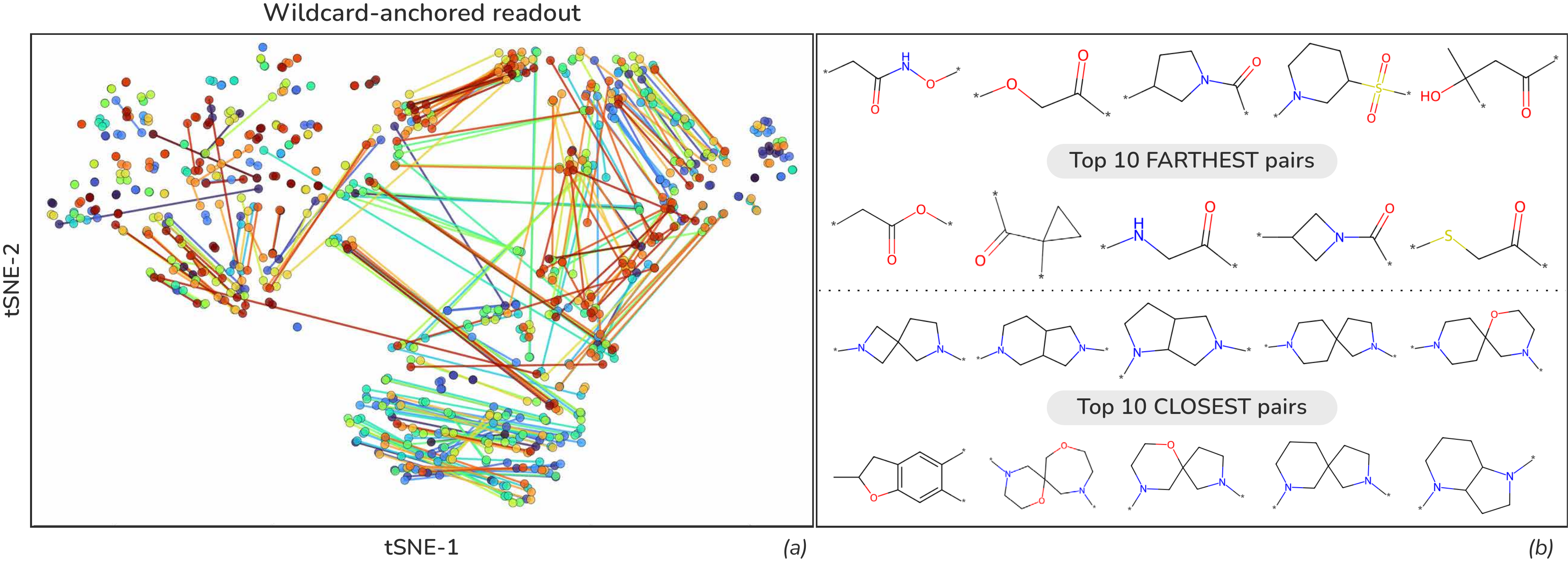}
  \caption{\textit{(a)} tSNE visualization of wildcard-anchored readouts. Each fragment contributes two readouts, one per wildcard; paired readouts are linked and shown with the same color.
\textit{(b)} fragments with the largest and smallest distances between their two wildcard-anchored readouts. Farthest pairs have distinct attachment environments, whereas closest pairs are nearly symmetric.}
  \label{fig:pairs_tsne}
\vspace{-0.8em}
  
\end{figure}

The wildcard anchor is motivated by a simple hypothesis: for a fragment with multiple attachment sites, the readout should depend on the active site rather than only on the fragment identity. We test this directly using 400 asymmetric fragments with exactly two wildcards, which account for 40.5\% of the 1000-fragment vocabulary. For each fragment, we compute two readouts, one using each wildcard as the query. If the wildcard-anchored readout ignores the choice of attachment site, the two embeddings should be identical. If it captures attachment-specific context, the embeddings should differ. Figure~\ref{fig:pairs_tsne}(a) visualizes the readouts with t-SNE, linking the two readouts of each fragment and coloring them alike.
Many paired readouts are well separated, indicating that changing the active wildcard substantially alters the embedding. At the same time, the links are frequently near-parallel, suggesting that switching the active wildcard induces a consistent transformation in the learned embedding space rather than an arbitrary perturbation.
Figure~\ref{fig:pairs_tsne}(b) makes this concrete: the fragments whose two readouts are most separated are those whose attachment sites sit in clearly different local environments, while the fragments whose readouts nearly coincide are close to symmetric.
The readout is therefore not a noisy copy of a global fragment embedding; it carries a site-specific view of the same chemistry, separating attachment contexts precisely when the geometry warrants it and collapsing them when it does not.

\vspace{-0.6em}
\subsubsection{Property-Conditional Generation Performance}
\vspace{-0.4em}
\label{subsec:conditional-generation-results}

We first evaluate property-conditional molecule generation against retrained baselines, then study how Fraglingo scales with training corpus and retrieval-vocabulary size.

\begin{table*}[!hbt]
  \centering
  \caption{Property-conditional molecule generation benchmarking results. Each model is followed by a row reporting the standard deviation over 3 runs where available. Best results are shown in \textbf{bold}; second-best results are \underline{underlined}.}
\vspace{-0.6em}
  \setlength{\tabcolsep}{3pt}
  \resizebox{\linewidth}{!}{
  \begin{tabular}{l|cccc|cccc|cccccccc}
    \toprule
    \multirow{3}{*}{\textbf{Model}} &
    \multicolumn{4}{c|}{\textbf{Generation Quality}} &
    \multicolumn{4}{c|}{\textbf{Joint Control}} &
    \multicolumn{8}{c}{\textbf{Per-Property Spearman}} \\
    \cmidrule(lr){2-5}
    \cmidrule(lr){6-9}
    \cmidrule(l){10-17}
    & \textbf{Validity}
    & \textbf{Uniqueness}
    & \textbf{Novelty}
    & \textbf{Diversity}
    & \textbf{NJD}
    & \textbf{Joint@1$\times$}
    & \textbf{Joint@2$\times$}
    & \textbf{Partial (/7)}
    & \textbf{logP}
    & \textbf{MW}
    & \textbf{QED}
    & \textbf{TPSA}
    & \textbf{HBD}
    & \textbf{HBA}
    & \textbf{RotBonds}
    & \textbf{Avg.} \\
    & ($\uparrow$)
    & ($\uparrow$)
    & ($\uparrow$)
    & ($\uparrow$)
    & ($\downarrow$)
    & ($\uparrow$)
    & ($\uparrow$)
    & ($\uparrow$)
    & ($\uparrow$)
    & ($\uparrow$)
    & ($\uparrow$)
    & ($\uparrow$)
    & ($\uparrow$)
    & ($\uparrow$)
    & ($\uparrow$)
    & ($\uparrow$)\\
    \midrule
    MolGPT
      & 25.7 & \textbf{100.0} & \textbf{100.0} & 0.3099
      & 1.7339 & \underline{53.97} & \underline{95.02} & \underline{6.32}
      & 0.8072 & \textbf{0.9442} & \underline{0.6659} & \textbf{0.8537} & 0.7340 & \underline{0.8242} & 0.6421 & \underline{0.7816} \\
    \quad (Std)
      & (1.9) & (0) & (0) & (0.026)
      & (0.113) & (1.362) & (3.945) & (0.190)
      & (0.038) & (0.031) & (0.074) & (0.037) & (0.057) & (0.039) & (0.056) & (0.012) \\
    MolRWKV
      & 90.9 & \textbf{100.0} & \textbf{100.0} & 0.8261
      & \underline{1.6090} & 43.77 & 90.46 & 6.17
      & \underline{0.8606} & 0.4830 & \textbf{0.7070} & 0.6449 & \textbf{0.8839} & \textbf{0.8572} & \textbf{0.8232} & 0.7514 \\
    \quad (Std)
      & (2.2) & (0) & (0) & (0.027)
      & (0.103) & (4.851) & (4.196) & (0.161)
      & (0.032) & (0.044) & (0.075) & (0.040) & (0.041) & (0.030) & (0.023) & (0.027) \\
    PS-VAE
      & \textbf{100.0} & 99.8 & \textbf{100.0} & \textbf{0.9029}
      & 2.6985 & 1.33 & 22.82 & 3.55
      & 0.4732 & 0.4774 & 0.0930 & 0.4100 & 0.1093 & 0.5485 & 0.1612 & 0.3246 \\
    \quad (Std)
      & (0) & (0.1) & (0) & (0.029)
      & (0.118) & (0.658) & (1.861) & (0.174)
      & (0.087) & (0.111) & (0.183) & (0.058) & (0.061) & (0.051) & (0.036) & (0.063) \\
    \textbf{Fraglingo (BRICS)}
      & \textbf{100.0} & \textbf{100.0} & \textbf{100.0} & 0.8444
      & 1.9808 & 40.88 & 85.90 & 5.80
      & 0.8373 & 0.6423 & 0.5632 & 0.7605 & 0.8161 & 0.7800 & 0.7760 & 0.7393 \\
    \quad (Std)
      & (0) & (0) & (0) & (0.027)
      & (0.054) & (2.111) & (3.602) & (0.191)
      & (0.035) & (0.033) & (0.042) & (0.066) & (0.046) & (0.081) & (0.043) & (0.038) \\
    \textbf{Fraglingo ablation}$^\dagger$
      & \textbf{100.0} & 99.8 & \textbf{100.0} & \underline{0.8478}
      & 2.1256 & 34.66 & 81.90 & 5.63
      & 0.7918 & 0.5128 & 0.4247 & 0.7254 & 0.8185 & 0.7377 & 0.6769 & 0.6697 \\
    \quad (Std)
      & (0) & (0.8) & (0) & (0.028)
      & (0.060) & (2.34) & (3.81) & (0.204)
      & (0.037) & (0.036) & (0.045) & (0.064) & (0.047) & (0.079) & (0.045) & (0.041) \\
    \textbf{Fraglingo (BFE)}
      & \textbf{100.0} & \underline{99.9} & \textbf{100.0} & 0.8218
      & \textbf{1.4928} & \textbf{69.65} & \textbf{97.75} & \textbf{6.56}
      & \underline{0.8710} & \underline{0.7746} & 0.6373 & \underline{0.8296} & \underline{0.8284} & 0.8190 & \underline{0.7993} & \textbf{0.7942} \\
    \bottomrule
  \end{tabular}
  }
  \vspace{2pt}
  \scriptsize{$^\dagger$ Wildcard-anchored readout ablated.}
  \label{tab:main_results}
\vspace{-2em}
\end{table*}

\begin{figure}[h]
  \centering
  \includegraphics[width=0.9\textwidth]{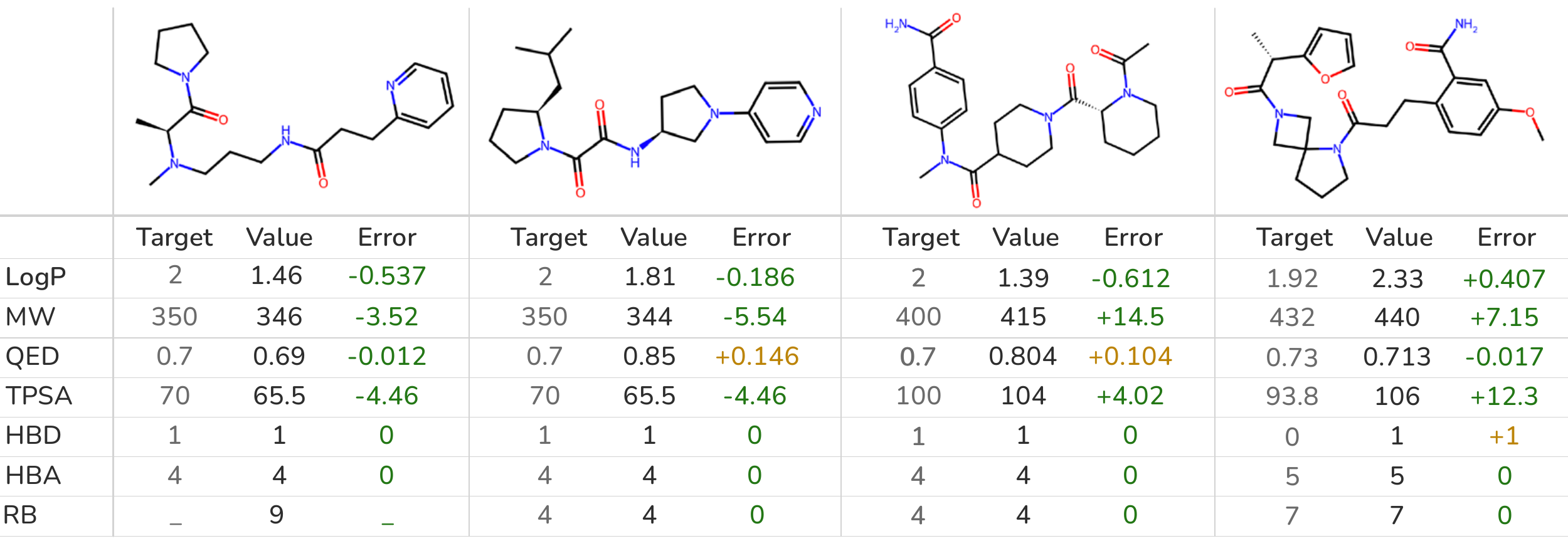}
  \caption{Examples from property-conditional \textbf{molecule generation}. Green marks properties close to the requested target, and yellow marks properties within the acceptable tolerance range.}
\vspace{-0.8em}
  \label{fig:eg_mol_gen}
  
\end{figure}

The difficulty of seven-way conditioning lies in satisfying all target properties simultaneously. Table~\ref{tab:main_results} shows that Fraglingo with BFE fragmentation~\citep{nguyen2026mollingo} gives the strongest joint control, with the lowest NJD (1.49), highest Joint@$1\times$ (69.65\%), highest Joint@$2\times$ (97.75\%), and highest average number of satisfied properties (6.56/7). Among the plain-BRICS settings, Fraglingo maintains 100\% validity, 100\% uniqueness, and 100\% novelty while the wildcard-anchored readout ablation degrades joint control (NJD 2.1256 and Joint@$1\times$ 34.66\%), supporting the role of the site-specific readout. MolGPT and MolRWKV are competitive on several aggregate metrics, whereas PS-VAE performs substantially worse under this stringent seven-property conditioning.
Figure~\ref{fig:eg_mol_gen} gives representative conditional generation samples. The examples illustrate the model can satisfy multiple property targets while producing chemically varied structures rather than repeatedly returning the same fragment pattern.

Table~\ref{tab:vocab_train_scaling} separates corpus scale from retrieval-library size, and the two act differently. At 1k molecules, the model remains data-limited: the best setting reaches NJD 3.93 and Joint@$1\times$ 12.24, while larger vocabularies mainly increase diversity. Increasing the corpus to 10k sharply improves property control, with the 100-fragment model reaching NJD 1.85, Joint@$1\times$ 45.83, and Partial 5.95/7. At 100k molecules, the 1000-fragment model gives the strongest joint control, with NJD 1.52, Joint@$1\times$ 58.98, Joint@$2\times$ 96.45, and Partial 6.29/7, while the 500-fragment model achieves the highest average per-property correlation (0.8052). Scaling to 1M molecules does not uniformly improve over 100k; its best joint-control setting uses 500 fragments and reaches NJD 1.57 and Joint@$1\times$ 57.18. Overall, larger vocabularies consistently increase diversity, but property control is strongest when corpus scale and retrieval-library size are balanced.
\begin{table*}[!hbt]
  \centering
  \caption{Property-conditional molecule generation under different vocabulary sizes and training dataset sizes.
\vspace{-0.6em}
  Best results are shown in \textbf{bold}; second-best results are \underline{underlined}.}
  \setlength{\tabcolsep}{3pt}
  \resizebox{\linewidth}{!}{
  \begin{tabular}{cc|cccc|cccc|cccccccc}
    \toprule
    \multirow{3}{*}{\textbf{Train}} &
    \multirow{3}{*}{\textbf{Vocab}} &
    \multicolumn{4}{c|}{\textbf{Generation Quality}} &
    \multicolumn{4}{c|}{\textbf{Joint Control}} &
    \multicolumn{8}{c}{\textbf{Per-Property Spearman}} \\
    \cmidrule(lr){3-6}
    \cmidrule(lr){7-10}
    \cmidrule(l){11-18}
    &
    & \textbf{Validity}
    & \textbf{Uniqueness}
    & \textbf{Novelty}
    & \textbf{Diversity}
    & \textbf{NJD}
    & \textbf{Joint@1$\times$}
    & \textbf{Joint@2$\times$}
    & \textbf{Partial (/7)}
    & \textbf{logP}
    & \textbf{MW}
    & \textbf{QED}
    & \textbf{TPSA}
    & \textbf{HBD}
    & \textbf{HBA}
    & \textbf{RotBonds}
    & \textbf{Avg.} \\
    &
    & ($\uparrow$)
    & ($\uparrow$)
    & ($\uparrow$)
    & ($\uparrow$)
    & ($\downarrow$)
    & ($\uparrow$)
    & ($\uparrow$)
    & ($\uparrow$)
    & ($\uparrow$)
    & ($\uparrow$)
    & ($\uparrow$)
    & ($\uparrow$)
    & ($\uparrow$)
    & ($\uparrow$)
    & ($\uparrow$)
    & ($\uparrow$) \\
    \midrule

    \multirow{4}{*}{1K}
    & 100  & 97.8 & 98.7 & \textbf{100.0} & 0.8344 & \textbf{3.9288} & \textbf{12.24} & \textbf{42.58} & \textbf{3.99} & \underline{0.5940} & \textbf{0.2953} & \underline{0.2920} & 0.3903 & \textbf{0.5228} & 0.3101 & \underline{0.4275} & \underline{0.4045} \\
    & 500  & 97.8 & 99.0 & \textbf{100.0} & 0.8631 & 4.2554 & \underline{11.84} & \underline{39.05} & \underline{3.82} & \textbf{0.6623} & \underline{0.1473} & \textbf{0.3138} & \textbf{0.4428} & \underline{0.4705} & \underline{0.3381} & \textbf{0.4717} & \textbf{0.4066} \\
    & 1000 & \underline{99.6} & \underline{99.3} & \textbf{100.0} & \underline{0.8748} & \underline{4.1446} & 10.10 & 36.15 & 3.78 & 0.4501 & 0.1098 & 0.2547 & \underline{0.4414} & 0.4598 & \textbf{0.3536} & 0.1643 & 0.3190 \\
    & 2000 & \textbf{99.8} & \textbf{99.6} & \textbf{100.0} & \textbf{0.8775} & 4.3054 & 11.25 & 35.18 & 3.78 & 0.5069 & 0.1394 & 0.2646 & 0.3582 & 0.4166 & 0.2783 & 0.4005 & 0.3377 \\
    \midrule

    \multirow{4}{*}{10K}
    & 100  & \textbf{100.0} & \textbf{100.0} & \textbf{100.0} & 0.8241 & \textbf{1.8532} & \textbf{45.83} & \textbf{91.00} & \textbf{5.95} & 0.7928 & \textbf{0.7242} & \textbf{0.6436} & \underline{0.8212} & 0.7287 & \textbf{0.8492} & \textbf{0.8188} & \textbf{0.7683} \\
    & 500  & \textbf{100.0} & \textbf{100.0} & \textbf{100.0} & 0.8444 & 1.9808 & 40.88 & 85.90 & \underline{5.80} & \underline{0.8373} & \underline{0.6423} & \underline{0.5632} & 0.7605 & \textbf{0.8161} & 0.7800 & \underline{0.7760} & \underline{0.7393} \\
    & 1000 & \textbf{100.0} & \underline{99.9} & \textbf{100.0} & \underline{0.8484} & \underline{1.8920} & \underline{44.13} & \underline{87.75} & \textbf{5.95} & 0.7655 & 0.5963 & 0.5495 & \textbf{0.8245} & \underline{0.8059} & \underline{0.7987} & 0.7738 & 0.7305 \\
    & 2000 & \textbf{100.0} & \textbf{100.0} & \textbf{100.0} & \textbf{0.8605} & 2.0198 & 38.28 & 86.25 & 5.73 & \textbf{0.8618} & 0.6016 & 0.5292 & 0.7510 & 0.7924 & 0.7472 & 0.7023 & 0.7122 \\
    \midrule

    \multirow{4}{*}{100K}
    & 100  & \textbf{100.0} & \textbf{100.0} & \textbf{100.0} & 0.8278 & 1.7492 & 51.83 & 91.15 & 6.06 & 0.7689 & 0.7362 & 0.5950 & 0.7323 & 0.6964 & 0.8063 & 0.7944 & 0.7327 \\
    & 500  & \textbf{100.0} & \textbf{100.0} & \textbf{100.0} & 0.8499 & \underline{1.5440} & \underline{56.68} & \underline{94.65} & \underline{6.22} & \textbf{0.8901} & \underline{0.7800} & 0.6361 & \textbf{0.8456} & 0.7890 & \textbf{0.8577} & \underline{0.8382} & \textbf{0.8052} \\
    & 1000 & \textbf{100.0} & \textbf{100.0} & \textbf{100.0} & \underline{0.8526} & \textbf{1.5189} & \textbf{58.98} & \textbf{96.45} & \textbf{6.29} & 0.8524 & 0.7501 & \textbf{0.6551} & 0.8391 & \textbf{0.8080} & 0.8038 & \textbf{0.8467} & 0.7935 \\
    & 2000 & \textbf{100.0} & \textbf{100.0} & \textbf{100.0} & \textbf{0.8616} & 1.6277 & 56.08 & 93.80 & 6.17 & \underline{0.8819} & \textbf{0.8026} & \underline{0.6500} & \underline{0.8442} & \underline{0.7981} & \underline{0.8191} & 0.8380 & \underline{0.8048} \\
    \midrule

    \multirow{4}{*}{1M}
    & 100  & \textbf{100.0} & \textbf{100.0} & \underline{99.8} & 0.8317 & 1.8422 & 52.10 & 88.79 & 5.98 & 0.7862 & 0.7703 & 0.6342 & 0.6829 & 0.6200 & 0.7767 & 0.7346 & 0.7149 \\
    & 500  & \textbf{100.0} & \textbf{100.0} & \textbf{100.0} & 0.8525 & \textbf{1.5668} & \textbf{57.18} & \textbf{94.25} & \textbf{6.20} & 0.8658 & \underline{0.8231} & 0.6249 & 0.7836 & 0.7710 & \underline{0.8259} & 0.8172 & 0.7873 \\
    & 1000 & \textbf{100.0} & \textbf{100.0} & \textbf{100.0} & \underline{0.8567} & \underline{1.6078} & 56.88 & 93.65 & \underline{6.19} & \textbf{0.8836} & 0.7691 & \textbf{0.6690} & \textbf{0.8012} & \textbf{0.7898} & \textbf{0.8298} & \underline{0.8280} & \textbf{0.7957} \\
    & 2000 & \textbf{100.0} & \underline{99.9} & \textbf{100.0} & \textbf{0.8651} & 1.6137 & \underline{57.16} & \underline{93.75} & 6.18 & \underline{0.8739} & \textbf{0.8546} & \underline{0.6451} & \underline{0.7953} & \underline{0.7718} & 0.7702 & \textbf{0.8450} & \underline{0.7936} \\
    \bottomrule
  \end{tabular}
  }
  \label{tab:vocab_train_scaling}
\vspace{-0.8em}
\end{table*}

\vspace{-0.6em}
\subsubsection{Continuous Retrieval Enables Vocabulary Expansion After Training}
\vspace{-0.4em}
\label{subsec:vocab-expansion}

The continuous-retrieval formulation enables a capability that classification-based generators cannot naturally support: because decoding is performed by nearest-neighbor search in a learned embedding space, the fragment library can be expanded after training without modifying or retraining the model.
We evaluate this capability directly by training with one fragment vocabulary and retrieving from progressively larger libraries at inference time.
Table~\ref{tab:cross_vocab_1000} demonstrates this capability. Expanding the inference library causes the model to select many fragments that were never present in the training vocabulary, while preserving saturated validity, uniqueness, and novelty. This expansion slightly increases molecular diversity, but it also weakens property control: NJD rises and the average number of satisfied property targets decreases as the retrieval library grows. The same pattern appears for both the 500-fragment and 1000-fragment training settings. These results show that inference-time vocabulary expansion provides a practical, training-free mechanism for exploring a larger chemical space. The strongest property control is achieved when the retrieval library matches the training distribution, allowing practitioners to trade a modest loss in accuracy for greater chemical coverage without retraining.
\begin{table*}[!hbt]
\vspace{-0.6em}
  \centering
  \caption{\textbf{Inference-time vocabulary expansion.} Models are trained with a fixed BFE fragment vocabulary but retrieve against a larger inference library at test time. OOV Frag.\% is the fraction of selected fragments absent from the training vocabulary.}
\vspace{-0.6em}
  \setlength{\tabcolsep}{3pt}
  \resizebox{\linewidth}{!}{
  \begin{tabular}{l|cccc|cc|ccccccc|c|c}
    \toprule
    \multirow{3}{*}{\textbf{Inference Vocabulary}} &
    \multicolumn{4}{c|}{\textbf{Generation Quality}} &
    \multicolumn{2}{c|}{\textbf{Joint Control}} &
    \multicolumn{7}{c|}{\textbf{Per-Property Spearman}} &
    \multirow{3}{*}{\textbf{Avg.}} &
    \multirow{3}{*}{\textbf{OOV Frag.}} \\
    \cmidrule(lr){2-5}
    \cmidrule(lr){6-7}
    \cmidrule(lr){8-14}
    & \textbf{Validity}
    & \textbf{Uniqueness}
    & \textbf{Novelty}
    & \textbf{Diversity}
    & \textbf{NJD}
    & \textbf{Partial (/7)}
    & \textbf{logP}
    & \textbf{MW}
    & \textbf{QED}
    & \textbf{TPSA}
    & \textbf{HBD}
    & \textbf{HBA}
    & \textbf{RotBonds}
    &
    & \\
    & ($\uparrow$)
    & ($\uparrow$)
    & ($\uparrow$)
    & ($\uparrow$)
    & ($\downarrow$)
    & ($\uparrow$)
    & ($\uparrow$)
    & ($\uparrow$)
    & ($\uparrow$)
    & ($\uparrow$)
    & ($\uparrow$)
    & ($\uparrow$)
    & ($\uparrow$)
    & ($\uparrow$)
    & \\
    \midrule
    Training (500-frags)
      & \textbf{100.0}
      & \underline{99.9}
      & \textbf{100.0}
      & 0.8218
      & \underline{1.4928}
      & \underline{6.56}
      & \textbf{0.8710}
      & 0.7746
      & \textbf{0.6373}
      & 0.8296
      & 0.8284
      & \underline{0.8190}
      & \textbf{0.7993}
      & \underline{0.7942}
      & \textbf{0.00\%} \\
    Inference (1000-frags)
      & \textbf{100.0}
      & \textbf{100.0}
      & \textbf{100.0}
      & 0.8313
      & 1.6028
      & 6.40
      & 0.8600
      & 0.7496
      & \underline{0.6323}
      & \textbf{0.8476}
      & \textbf{0.8534}
      & \textbf{0.8280}
      & \underline{0.7903}
      & \textbf{0.7945}
      & 29.11\% \\
    Inference (2000-frags)
      & \textbf{100.0}
      & \textbf{100.0}
      & \textbf{100.0}
      & \textbf{0.8382}
      & 1.8628
      & 6.16
      & \underline{0.8620}
      & 0.7286
      & 0.5953
      & 0.8226
      & \underline{0.8334}
      & 0.8150
      & 0.7773
      & 0.7763
      & 48.76\% \\

    \midrule
    Training (1000-frags)
      & \textbf{100.0}
      & \textbf{100.0}
      & \textbf{100.0}
      & 0.8313
      & \textbf{1.2318}
      & \textbf{6.83}
      & 0.7740
      & \textbf{0.8896}
      & 0.6043
      & \underline{0.8456}
      & 0.7354
      & 0.7890
      & 0.7523
      & 0.7702
      & \textbf{0.00\%} \\
    Inference (2000-frags)
      & \textbf{100.0}
      & \textbf{100.0}
      & \textbf{100.0}
      & \underline{0.8362}
      & 1.5018
      & 6.54
      & 0.7560
      & \underline{0.8676}
      & 0.6043
      & 0.8376
      & 0.7374
      & 0.7855
      & 0.7653
      & 0.7648
      & \underline{28.32\%} \\
    \bottomrule
  \end{tabular}
  }
  \label{tab:cross_vocab_1000}
\vspace{-0.8em}
  
\end{table*}

\vspace{-0.8em}
\section{Related Work}
\vspace{-0.8em}
\label{sec:related}
Early molecular generators represent molecules as SMILES strings or molecular graphs. SMILES-based methods use recurrent networks, variational autoencoders, or Transformers~\citep{segler2018generating,gomez2018automatic,ross2022large}, while graph-based methods generate atoms and bonds through autoregressive, flow, or diffusion processes~\citep{shi2020graphaf,hoogeboom2022equivariant}. Despite architectural differences, these approaches largely operate at the atom or token level, leaving chemically meaningful substructures to be learned implicitly. Fragment-based methods instead generate larger chemical units that more closely reflect medicinal-chemistry operations. Methods including HierVAE~\citep{jin2020hierarchical}, JT-VAE~\citep{jin2018junction}, PS-VAE~\citep{kong2022molecule}, MoLeR~\citep{maziarz2021learning}, and FragGPT~\citep{yue2024unlocking} generate discrete fragments from fixed vocabularies, while t-SMILES~\citep{wu2024t} and SynCoGen~\citep{rekesh2025syncogen} explore hierarchical and synthesis-oriented fragment representations. These methods nevertheless rely on discrete fragment or building-block vocabularies, limiting expansion to unseen fragments. Fraglingo instead uses attachment-aware continuous latent retrieval, allowing the fragment library to expand at inference time while conditioning retrieval on the active attachment site.
\vspace{-0.8em}
\section{Conclusion, Limitations, and Future Work}
\vspace{-0.6em}
\label{sec:conclusion}

Fraglingo introduces a retrieval-based formulation for fragment-level molecular design. Instead of classifying over fixed fragment identifiers, the model predicts a continuous fragment embedding and decodes it by nearest-neighbor retrieval. Combined with a wildcard-anchored readout that conditions retrieval on the active attachment site, this formulation jointly determines fragment identity and attachment while enabling inference-time expansion of the fragment vocabulary without retraining. Across molecule generation, scaffold generation, scaffold decoration, and molecule optimization, the same latent retrieval architecture is reused by changing only the retrieval query and conditioning signal.
Fraglingo remains limited by the chosen fragmentation scheme and vocabulary. Future work includes richer fragmentation strategies, stronger conditioning mechanisms, and scaling retrieval to substantially larger fragment libraries.

\subsubsection*{Acknowledgments}
This work was supported by the NSF Molecule Maker Lab Institute (MMLI), an AI Institute for Molecular Discovery, Synthesis Strategy, and Manufacturing, funded by the U.S. National Science Foundation under Awards No. 2019897 and 2505932.

\clearpage
\bibliography{iclr2026_conference}
\bibliographystyle{iclr2026_conference}

\clearpage
\appendix
\section{Ablations Attribute the Gains to Continuous Retrieval}
\label{appendix:component-ablation}

Section~\ref{subsec:conditional-generation-results} shows that Fraglingo achieves strong property control. Here, we isolate the contribution of each design component by holding the setting fixed (10k corpus, 500 fragments, 6.4M parameters) and changing one component at a time:

\begin{itemize}[leftmargin=*]
    \item \textbf{Contrastive vs.\ regression objective.} Replace InfoNCE with a plain regression loss ($\|\hat{z}-e_2\|^2$) to the true fragment embedding, keeping nearest-neighbor decoding. This asks whether the retrieval geometry needs in-batch negatives or only a target direction.
    \item \textbf{Classification loss.} Remove $\mathcal{L}_{\text{cls}}$ to test the embedding-collapse hypothesis of Section~\ref{sec:fraggen}: without a discriminative pressure on fragment embeddings, InfoNCE alone may map distinct fragments to overlapping regions.
    \item \textbf{RVQ regularizer.} Remove $\mathcal{L}_{\text{commit}}$ and the codebook to test whether RVQ further organizes the latent space beyond the classification objective, encouraging compact and retrievable fragment embeddings.
\end{itemize}

\begin{table*}[!hbt]
  \centering
  \caption{Component ablation on the shared 10k-molecule, 500-fragment setting. Each row removes or replaces exactly one component relative to the full model (top row); rows are ordered by severity of the resulting drop in joint control.}
  \setlength{\tabcolsep}{4pt}
  \resizebox{\linewidth}{!}{
  \begin{tabular}{l|cccc|cccc|cccccccc}
    \toprule
    \multirow{3}{*}{\textbf{Configuration}} &
    \multicolumn{4}{c|}{\textbf{Generation Quality}} &
    \multicolumn{4}{c|}{\textbf{Joint Control}} &
    \multicolumn{8}{c}{\textbf{Per-Property Spearman}} \\
    \cmidrule(lr){2-5}
    \cmidrule(lr){6-9}
    \cmidrule(l){10-17}
    & \textbf{Validity} & \textbf{Uniqueness} & \textbf{Novelty} & \textbf{Diversity}
    & \textbf{NJD} & \textbf{Joint@1$\times$} & \textbf{Joint@2$\times$} & \textbf{Partial (/7)}
    & \textbf{logP} & \textbf{MW} & \textbf{QED} & \textbf{TPSA} & \textbf{HBD} & \textbf{HBA} & \textbf{RotBonds} & \textbf{Avg.} \\
    & ($\uparrow$) & ($\uparrow$) & ($\uparrow$) & ($\uparrow$)
    & ($\downarrow$) & ($\uparrow$) & ($\uparrow$) & ($\uparrow$)
    & ($\uparrow$) & ($\uparrow$) & ($\uparrow$) & ($\uparrow$) & ($\uparrow$) & ($\uparrow$) & ($\uparrow$) & ($\uparrow$) \\
    \midrule
    Full model (Fraglingo)
      & \textbf{100.0} & \underline{99.9} & \textbf{100.0} & 0.8218
      & \textbf{1.4928} & \textbf{69.65} & \textbf{97.75} & \textbf{6.56}
      & \textbf{0.8710} & \underline{0.7746} & \textbf{0.6373} & \textbf{0.8296} & \textbf{0.8284} & \textbf{0.8190} & \textbf{0.7993} & \textbf{0.7942} \\
    \midrule
    InfoNCE ($\to$ MSE regression)
      & 99.8 & \textbf{100.0} & \textbf{100.0} & \underline{0.8238}
      & 4.2248 & 26.85 & 24.94 & 3.73
      & 0.1010 & 0.3146 & 0.0033 & 0.3636 & 0.1674 & 0.1470 & 0.1453 & 0.1772 \\
    Classification loss
      & \textbf{100.0} & \textbf{100.0} & \textbf{100.0} & 0.8228
      & 1.7878 & 56.52 & 88.14 & 5.95
      & 0.7570 & 0.6966 & 0.4823 & 0.7756 & \underline{0.7434} & 0.7520 & 0.7193 & 0.7032 \\
    RVQ regularizer
      & \textbf{100.0} & \textbf{100.0} & \textbf{100.0} & \textbf{0.8268}
      & \underline{1.5268} & \underline{67.75} & \underline{95.76} & \underline{6.29}
      & \underline{0.7780} & \textbf{0.8536} & \underline{0.6053} & \underline{0.8046} & 0.7344 & \underline{0.7980} & \underline{0.7243} & \underline{0.7562} \\
    \bottomrule
  \end{tabular}
  }
  \label{tab:component_ablation}
\end{table*}

The ablation cleanly attributes the model's control to the continuous-retrieval objective. Replacing InfoNCE with direct MSE regression is catastrophic: NJD rises from 1.4928 to 4.2248, Joint@$1\times$ drops from 69.65\% to 26.85\%, and average per-property correlation falls from 0.7942 to 0.1772, essentially destroying property control while generation quality stays intact. Regressing to a target point is not enough; the contrastive objective, with its in-batch negatives, is what organizes the embedding space into a geometry where nearest-neighbor retrieval selects the right fragment. This is the core mechanism behind the first contribution, and the ablation shows it is not optional but load-bearing.

The two regularizers play supporting roles consistent with their design intent. Removing the classification loss degrades control (NJD 1.7878, Joint@$1\times$ 56.52\%, Avg. 0.7032), confirming the embedding-collapse hypothesis: without a discriminative pressure, distinct fragments blur together and retrieval loses precision. Removing RVQ has the mildest effect (NJD 1.5268, Joint@$1\times$ 67.75\%), so RVQ is a light stabilizer rather than a primary signal; we keep it because it costs little and slightly improves control, but the method does not depend on it.

\section{One Primitive Transfers to Molecule Optimization}
\label{1:molecule-optimization-results}

The next three sections evaluate Fraglingo beyond molecule generation, showing how the retrieval mechanism extends to molecule optimization, scaffold generation, and scaffold decoration by changing only the query and conditioning signals. We begin with molecule optimization.

\begin{table*}[!hbt]
  \centering
  \caption{Molecule optimization against fragment-based baselines on the shared 10k matched-pair dataset. Tanimoto and Sim$\ge$0.6 measure preservation of the starting molecule.}
  \setlength{\tabcolsep}{3pt}
  \resizebox{\linewidth}{!}{
  \begin{tabular}{l|cc|cccc|cccccccc}
    \toprule
    \multirow{3}{*}{\textbf{Model}} &
    \multicolumn{2}{c|}{\textbf{Generation Quality}} &
    \multicolumn{4}{c|}{\textbf{Optimization Performance}} &
    \multicolumn{8}{c}{\textbf{Property Distribution Similarity}} \\
    \cmidrule(lr){2-3}
    \cmidrule(lr){4-7}
    \cmidrule(l){8-15}
    & \textbf{Validity}
    & \textbf{Novelty}
    & \textbf{NJD}
    & \textbf{Tanimoto}
    & \textbf{Sim$\ge$0.6}
    & \textbf{Partial (/7)}
    & \textbf{logP}
    & \textbf{MW}
    & \textbf{QED}
    & \textbf{TPSA}
    & \textbf{HBD}
    & \textbf{HBA}
    & \textbf{RotBonds}
    & \textbf{Avg.} \\
    & ($\uparrow$)
    & ($\uparrow$)
    & ($\downarrow$)
    & ($\uparrow$)
    & ($\uparrow$)
    & ($\uparrow$)
    & ($\uparrow$)
    & ($\uparrow$)
    & ($\uparrow$)
    & ($\uparrow$)
    & ($\uparrow$)
    & ($\uparrow$)
    & ($\uparrow$)
    & ($\uparrow$) \\
    \midrule
    HierVAE
      & \textbf{100.0\%}
      & \textbf{100.0\%}
      & \underline{1.7717}
      & \underline{0.6746}
      & \underline{66.14\%}
      & \textbf{6.17}
      & \underline{0.821}
      & \underline{0.723}
      & \textbf{0.727}
      & \underline{0.719}
      & \underline{0.817}
      & \underline{0.809}
      & \underline{0.719}
      & \underline{0.762} \\
    \quad (Std)
      & (0)
      & (0)
      & (0.011)
      & (0.014)
      & (2.6)
      & (0.006)
      & (0.010)
      & (0.012)
      & (0.009)
      & (0.010)
      & (0.014)
      & (0.004)
      & (0.004)
      & (0.011) \\
    JT-VAE
      & \textbf{100.0\%}
      & \underline{99.9\%}
      & 2.6481
      & 0.2940
      & 4.12\%
      & 5.19
      & 0.509
      & 0.329
      & 0.373
      & 0.277
      & 0.537
      & 0.421
      & 0.515
      & 0.423 \\
    \quad (Std)
      & (0)
      & (0)
      & (0.093)
      & (0.027)
      & (0.74)
      & (0.60)
      & (0.078)
      & (0.069)
      & (0.090)
      & (0.063)
      & (0.093)
      & (0.071)
      & (0.102)
      & (0.098) \\
    \textbf{Fraglingo}
      & \textbf{100.0\%}
      & \textbf{100.0\%}
      & \textbf{1.6244}
      & \textbf{0.7952}
      & \textbf{95.72\%}
      & \textbf{6.17}
      & \textbf{0.839}
      & \textbf{0.744}
      & \underline{0.713}
      & \textbf{0.772}
      & \textbf{0.869}
      & \textbf{0.840}
      & \textbf{0.788}
      & \textbf{0.795} \\
    \quad (Std)
      & (0)
      & (0)
      & (0.018)
      & (0.008)
      & (1.0)
      & (0.010)
      & (0.004)
      & (0.007)
      & (0.004)
      & (0.014)
      & (0.011)
      & (0.003)
      & (0.006)
      & (0.009) \\
    \bottomrule
  \end{tabular}
  }
  \label{tab:optimization_comparison}
\end{table*}

We benchmark our Fraglingo's molecule optimization model against HierVAE and JT-VAE under the shared matched-pair optimization protocol. All optimization models are trained on the same 10k matched-pair dataset; fragment-based models use a vocabulary of 500 fragments. As described in Section~\ref{sec:optimization}, reference-molecule conditioning turns Fraglingo's attachment-conditioned retrieval mechanism into a molecule optimizer; here it gives the strongest target matching among the compared fragment-based methods. Table~\ref{tab:optimization_comparison} shows Fraglingo attains the lowest final NJD (1.6244), preserves the starting molecule best among the compared methods (Tanimoto 0.7952 and 95.72\% above 0.6 similarity), and gives the best average distributional similarity. Fraglingo makes similarity-constrained edits when needed to hit multi-property targets, which is the intended behavior for property-directed optimization.

\begin{figure}[h]
  \centering
  \includegraphics[width=1\textwidth]{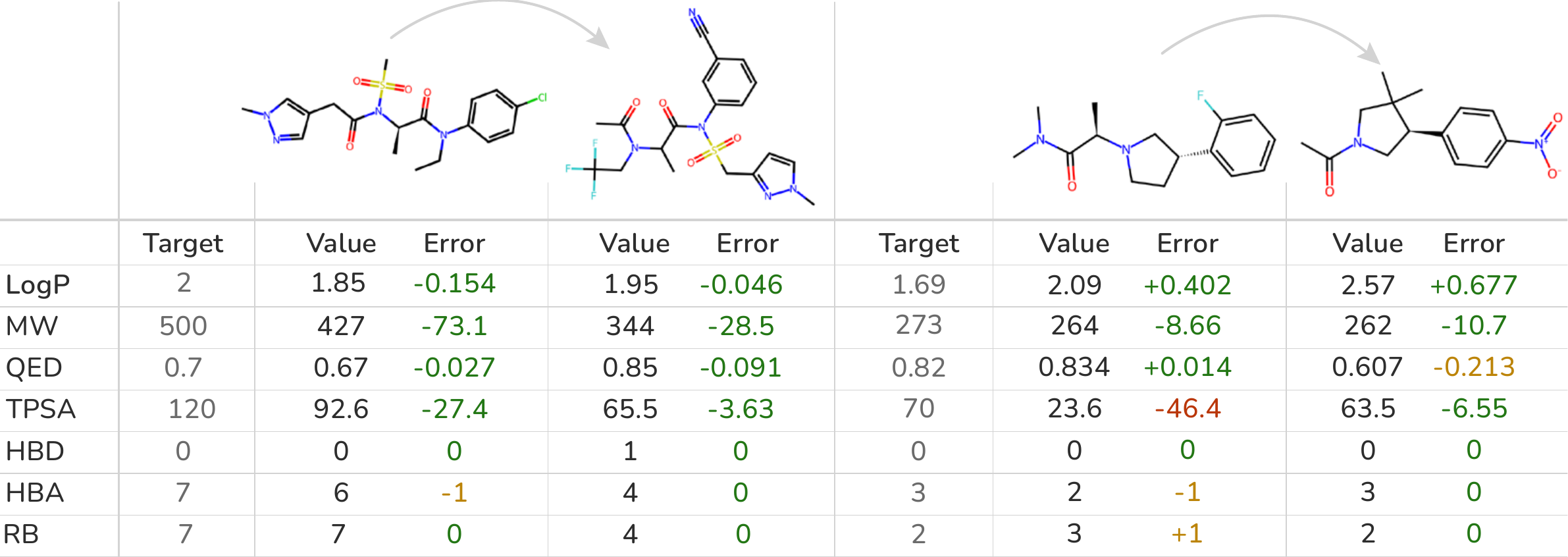}
  \caption{Qualitative \textbf{molecule optimization} examples. Green marks properties close to target, yellow marks properties within tolerance, and red marks properties outside the target range.}
  \label{fig:eg_mol_opt}
  
\end{figure}

Figure~\ref{fig:eg_mol_opt} shows representative optimization examples produced by Fraglingo. The examples highlight the intended edit behavior: Fraglingo changes local fragments to move properties toward the target while preserving much of the starting molecule.

\begin{table*}[!hbt]
  \centering
  \caption{Ablation on training corpus size and vocabulary size for molecule optimization. Tanimoto and Sim$\ge$0.6 measure preservation of the starting molecule.}
  \setlength{\tabcolsep}{3pt}
  \resizebox{\linewidth}{!}{
  \begin{tabular}{cc|c|cccc|cccccccc}
    \toprule
    \multirow{3}{*}{\textbf{Train}} &
    \multirow{3}{*}{\textbf{Vocab}} &
    \multicolumn{1}{c|}{\textbf{Generation}} &
    \multicolumn{4}{c|}{\textbf{Optimization Performance}} &
    \multicolumn{8}{c}{\textbf{Property Distribution Similarity}} \\
    \cmidrule(lr){3-3}
    \cmidrule(lr){4-7}
    \cmidrule(l){8-15}
    & &
    \textbf{Validity}
    & \textbf{NJD}
    & \textbf{Tanimoto}
    & \textbf{Sim$\ge$0.6}
    & \textbf{Partial (/7)}
    & \textbf{logP}
    & \textbf{MW}
    & \textbf{QED}
    & \textbf{TPSA}
    & \textbf{HBD}
    & \textbf{HBA}
    & \textbf{RotBonds}
    & \textbf{Avg.} \\
    & &
    ($\uparrow$)
    & ($\downarrow$)
    & ($\uparrow$)
    & ($\uparrow$)
    & ($\uparrow$)
    & ($\uparrow$)
    & ($\uparrow$)
    & ($\uparrow$)
    & ($\uparrow$)
    & ($\uparrow$)
    & ($\uparrow$)
    & ($\uparrow$)
    & ($\uparrow$) \\
    \midrule
    10k  & \multirow{3}{*}{500} & \textbf{100.0\%} & 1.6244 & \underline{0.7952} & \underline{95.72\%} & 6.17 & 0.8390 & 0.7440 & 0.7130 & 0.7720 & 0.8690 & 0.8400 & 0.7880 & 0.7950 \\
    100k &                       & \textbf{100.0\%} & 1.4031 & 0.7903 & \textbf{95.82\%} & \underline{6.34} & \underline{0.8709} & 0.7712 & 0.7681 & 0.8327 & 0.8753 & 0.8593 & 0.8103 & 0.8269 \\
    1M   &                       & \textbf{100.0\%} & \textbf{1.2037} & 0.7853 & 94.62\% & \textbf{6.39} & 0.8617 & \underline{0.7830} & \textbf{0.7981} & 0.8351 & \underline{0.8941} & \textbf{0.8927} & \textbf{0.8791} & \underline{0.8491} \\
    \midrule
    10k  & \multirow{3}{*}{1000} & \textbf{100.0\%} & 1.6477 & 0.7896 & 94.72\% & 6.12 & 0.8414 & 0.6757 & 0.7280 & 0.7772 & 0.8656 & 0.8328 & 0.7924 & 0.7876 \\
    100k &                        & \textbf{100.0\%} & 1.4429 & 0.7747 & 94.42\% & 6.24 & 0.8558 & 0.7664 & 0.7629 & 0.8146 & 0.8549 & 0.8227 & 0.8330 & 0.8158 \\
    1M   &                        & \textbf{100.0\%} & \underline{1.2820} & 0.7653 & 93.52\% & 6.30 & 0.8519 & \textbf{0.8127} & \underline{0.7867} & \textbf{0.8537} & 0.8906 & \underline{0.8881} & 0.8668 & \textbf{0.8501} \\
    \midrule
    10k  & \multirow{3}{*}{2000} & \textbf{100.0\%} & 1.6232 & \textbf{0.7980} & 95.02\% & 6.13 & 0.8273 & 0.7136 & 0.7478 & 0.7639 & 0.8731 & 0.8258 & 0.8426 & 0.7992 \\
    100k &                        & \textbf{100.0\%} & 1.4840 & 0.7774 & 93.12\% & 6.20 & 0.8241 & 0.7196 & 0.7489 & 0.8201 & 0.8788 & 0.8508 & 0.8360 & 0.8112 \\
    1M   &                        & \textbf{100.0\%} & 1.3082 & 0.7669 & 92.82\% & 6.31 & \textbf{0.8747} & 0.7695 & 0.7831 & \underline{0.8488} & \textbf{0.8965} & 0.8646 & \underline{0.8716} & 0.8441 \\
    \bottomrule
  \end{tabular}
  }
  \label{tab:opt_ablation}
\end{table*}

Table~\ref{tab:opt_ablation} shows that increasing the matched-pair training corpus consistently improves target matching, whereas the effect of vocabulary size is more nuanced. The 1M-pair, 500-fragment model achieves the lowest NJD (1.2037), while the 1M-pair, 1000-fragment model gives the highest average property-distribution similarity (0.8501). Larger vocabularies can improve preservation for smaller training sets---the 10k-pair, 2000-fragment model attains the highest Tanimoto similarity (0.7980)---but do not consistently improve target matching as the corpus grows. All configurations maintain 100\% validity, indicating that training-set coverage is the primary driver of optimization performance once the vocabulary provides sufficient edit diversity.

\section{\ldots{} to Scaffold Generation}
\label{2:scaffold-generation-results}

\begin{figure}[!hbt]
  \centering
  \includegraphics[width=\textwidth]{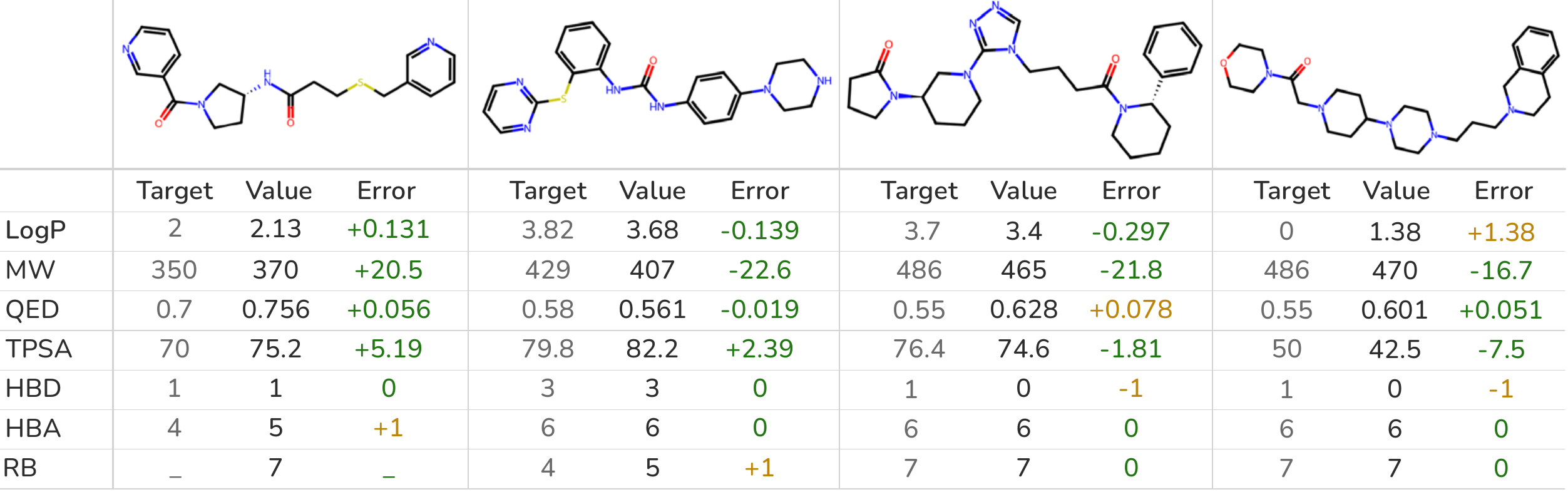}
  \caption{Qualitative \textbf{scaffold generation} examples. Fraglingo assembles scaffold-level BRICS fragments into valid core structures before side-chain decoration, producing diverse ring systems and attachment-ready topologies rather than complete decorated molecules.}
  \label{fig:eg_scf_gen}  
\end{figure}

We evaluate scaffold generation through controlled self-evaluation because, to the best of our knowledge, no available scaffold-generation baseline can be retrained under the same protocol. The evaluation therefore focuses on whether Fraglingo, when retargeted to scaffold fragments, produces valid ring systems and diverse core topologies (Figure~\ref{fig:eg_scf_gen}), and on how property control varies with vocabulary and corpus size (Table~\ref{tab:scaffold_ablation}). Scaffold generation can be viewed as a constrained version of full-molecule generation, with retrieval limited to scaffold fragments.

\begin{table*}[!hbt]
  \centering
  \caption{Scaffold generation under different vocabulary sizes and training dataset sizes.}
  \setlength{\tabcolsep}{3pt}
  \resizebox{\linewidth}{!}{
  \begin{tabular}{cc|cccc|cccc|cccccccc}
    \toprule
    \multirow{3}{*}{\textbf{Train}} &
    \multirow{3}{*}{\textbf{Vocab}} &
    \multicolumn{4}{c|}{\textbf{Generation Quality}} &
    \multicolumn{4}{c|}{\textbf{Joint Control}} &
    \multicolumn{8}{c}{\textbf{Per-Property Spearman}} \\
    \cmidrule(lr){3-6}
    \cmidrule(lr){7-10}
    \cmidrule(l){11-18}
    &
    & \textbf{Validity}
    & \textbf{Uniqueness}
    & \textbf{Novelty}
    & \textbf{Diversity}
    & \textbf{NJD}
    & \textbf{Joint@1$\times$}
    & \textbf{Joint@2$\times$}
    & \textbf{Partial (/7)}
    & \textbf{logP}
    & \textbf{MW}
    & \textbf{QED}
    & \textbf{TPSA}
    & \textbf{HBD}
    & \textbf{HBA}
    & \textbf{RotBonds}
    & \textbf{Avg.} \\
    &
    & ($\uparrow$)
    & ($\uparrow$)
    & ($\uparrow$)
    & ($\uparrow$)
    & ($\downarrow$)
    & ($\uparrow$)
    & ($\uparrow$)
    & ($\uparrow$)
    & ($\uparrow$)
    & ($\uparrow$)
    & ($\uparrow$)
    & ($\uparrow$)
    & ($\uparrow$)
    & ($\uparrow$)
    & ($\uparrow$)
    & ($\uparrow$) \\
    \midrule
    \multirow{4}{*}{10k}
      & 50  & \textbf{100.0} & 96.8 & 88.2 & 0.8059 & \underline{2.5943} & \underline{41.67} & \underline{84.64} & \underline{5.68} & 0.6169 & \textbf{0.7711} & \underline{0.4029} & 0.5954 & \underline{0.6196} & 0.5987 & 0.6701 & \underline{0.6107} \\
      & 100 & \textbf{100.0} & 97.0 & 89.4 & \underline{0.8262} & \textbf{2.5278} & \textbf{44.22} & \textbf{84.99} & \textbf{5.77} & \textbf{0.7397} & \underline{0.7302} & 0.3994 & \underline{0.6019} & 0.5859 & \underline{0.6052} & \textbf{0.7786} & \textbf{0.6344} \\
      & 200 & \textbf{100.0} & \underline{99.0} & \underline{95.2} & \underline{0.8363} & 3.1045 & 32.05 & 74.45 & 5.28 & \underline{0.7253} & 0.6606 & 0.2815 & 0.5352 & \textbf{0.6709} & 0.4859 & 0.5754 & 0.5621 \\
      & 500 & 99.9 & \textbf{99.2} & \textbf{97.0} & 0.8397 & 3.9841 & 25.28 & 63.91 & 4.74 & 0.6526 & 0.4954 & 0.1639 & 0.4121 & 0.6040 & 0.2993 & 0.6391 & 0.4666 \\
    \midrule
    \multirow{4}{*}{100k}
      & 50  & 99.9 & 98.8 & 82.8 & 0.8082 & 2.7067 & 37.04 & 78.83 & 5.48 & 0.5840 & 0.5958 & 0.3421 & 0.5226 & 0.4922 & 0.5371 & 0.5929 & 0.5238 \\
      & 100 & 99.9 & \underline{99.4} & 90.0 & 0.8219 & \underline{2.4809} & \underline{42.94} & \underline{81.58} & 5.65 & 0.6564 & 0.6445 & \textbf{0.4560} & \textbf{0.6031} & 0.6055 & \textbf{0.6137} & \underline{0.6474} & 0.6038 \\
      & 200 & 99.3 & 99.3 & \underline{93.6} & \textbf{0.8404} & 3.0413 & 33.79 & 71.50 & 5.22 & 0.6512 & 0.6276 & 0.3659 & 0.5720 & 0.5105 & 0.4341 & \underline{0.7090} & 0.5529 \\
      & 500 & 99.9 & \textbf{99.6} & \textbf{97.0} & \textbf{0.8464} & 3.4155 & 31.83 & 64.71 & 5.06 & 0.5277 & 0.4496 & 0.3100 & 0.5591 & 0.6119 & 0.4730 & 0.6463 & 0.5111 \\
    \midrule
    \multirow{4}{*}{1M}
      & 50  & \textbf{100.0} & 99.3 & 80.7 & 0.8097 & 2.4646 & 39.67 & \underline{83.19} & \underline{5.69} & 0.6037 & 0.6050 & 0.3359 & 0.5659 & 0.5024 & 0.5036 & 0.5694 & 0.5266 \\
      & 100 & \textbf{100.0} & 99.4 & 91.2 & 0.8141 & \textbf{2.1501} & \textbf{46.45} & \textbf{87.05} & \textbf{5.94} & \textbf{0.7783} & \textbf{0.6790} & \textbf{0.4996} & \textbf{0.6379} & 0.5802 & \textbf{0.6558} & \underline{0.7109} & \textbf{0.6488} \\
      & 200 & \textbf{100.0} & \textbf{99.8} & \underline{93.6} & \underline{0.8333} & \underline{2.4190} & \underline{41.62} & 81.79 & 5.65 & \underline{0.6885} & \underline{0.6271} & \underline{0.4551} & \underline{0.5838} & \underline{0.5940} & \underline{0.5915} & \underline{0.7098} & \underline{0.6071} \\
      & 500 & \underline{99.9} & \underline{99.7} & \textbf{95.6} & \textbf{0.8441} & 2.9393 & 33.88 & 70.67 & 5.29 & 0.6526 & 0.5193 & 0.3484 & 0.5514 & \textbf{0.6140} & 0.5585 & \textbf{0.7671} & 0.5730 \\
    \bottomrule
  \end{tabular}
  }
  \label{tab:scaffold_ablation}
\end{table*}

Table~\ref{tab:scaffold_ablation} shows a stronger vocabulary-size tradeoff for scaffold-only generation than for full-molecule generation, and this pattern remains visible at the 1M training-corpus scale. Small vocabularies, especially 100 fragments, give the best joint-control metrics because they concentrate probability mass on common scaffold motifs and make the retrieval problem easier. Larger vocabularies increase uniqueness, novelty, and diversity, but the additional rare fragments worsen NJD and joint satisfaction performance. This behavior is expected for scaffolds: the model must control global core topology with fewer substituent degrees of freedom, so an overly broad scaffold vocabulary can make property-conditioned retrieval less reliable.

\section{\ldots{} and to Scaffold Decoration}
\label{3:scaffold-decoration-results}

Scaffold decoration is again evaluated by self-comparison across settings rather than against external systems, for the same protocol-compatibility reason. Prompt-based decoration methods such as PromptSMILES~\citep{thomas2024promptsmiles} use different conditioning interfaces, while scaffold-extension methods such as MoLeR are primarily optimized for structural plausibility rather than explicit multi-property control. The question here is whether site-conditioned retrieval chooses chemically plausible branches for a fixed core while keeping scaffold atom indices stable across sites.

\begin{figure}[!hbt]
  \centering
  \includegraphics[width=\textwidth]{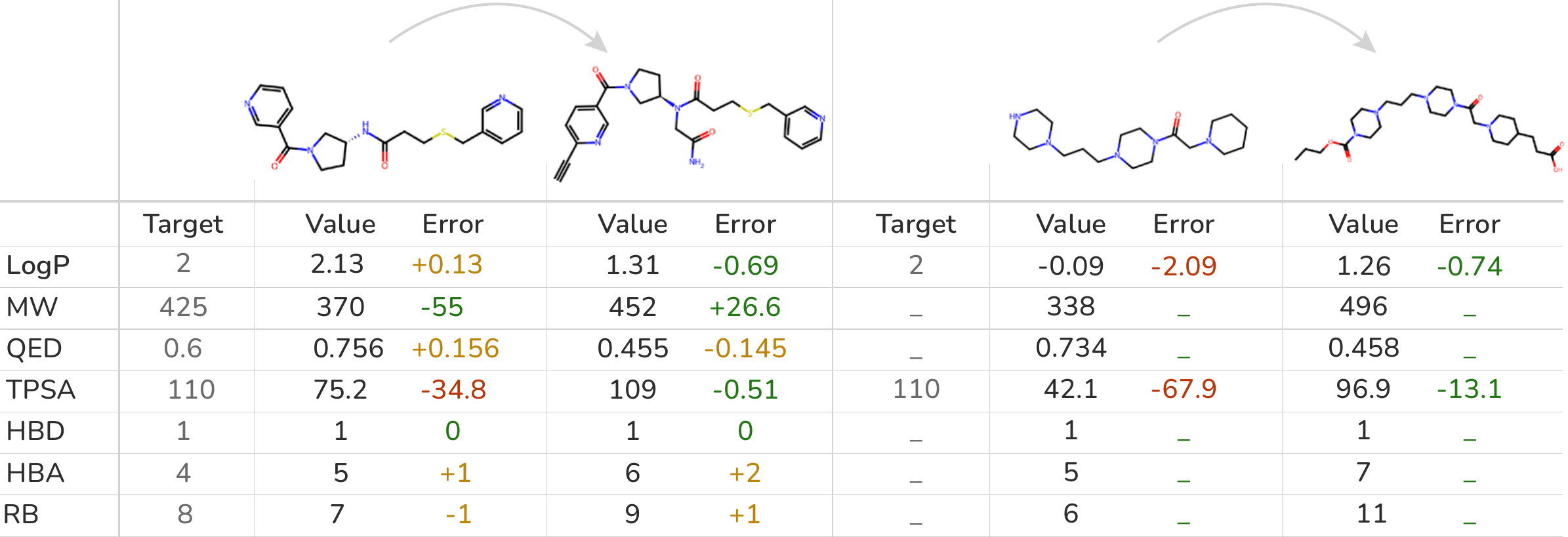}
  \caption{Qualitative \textbf{scaffold decoration} examples. For each fixed scaffold, Fraglingo predicts branch fragments at the growable sites under the specified property targets. The tables report molecular-property values and signed errors before and after decoration, illustrating how site-conditioned retrieval can move decorated molecules toward the requested property profile while preserving the scaffold core.}
  \label{fig:eg_scf_decor}
  
\end{figure}

Figure~\ref{fig:eg_scf_decor} illustrates the behavior of the decoration model on fixed scaffolds with explicit property targets. Because each attachment site is encoded from the viewpoint of the corresponding scaffold atom, the model selects different branch fragments for different local environments while leaving the core unchanged. Retrieved decorations reduce errors on targeted properties such as TPSA and molecular weight, though improving one property can trade off against weakly constrained descriptors. This matches the intended use of decoration as a controlled local edit: the scaffold fixes the conserved core, and Fraglingo searches the branch-embedding space for site-compatible substituents aligned with the requested profile.

Table~\ref{tab:decoration_results} quantifies this over the held-out decoration set. We report the same generation-quality and joint-control metrics as for molecule generation, computed on decorated molecules while the scaffold is held fixed.

\begin{table*}[!hbt]
  \centering
  \caption{\textbf{Scaffold decoration} under different vocabulary and corpus sizes. Metrics are computed on decorated molecules with the scaffold held fixed.}
  \setlength{\tabcolsep}{10pt}
  \resizebox{\linewidth}{!}{
  \begin{tabular}{cc|ccc|cccccccc}
    \toprule
    \multirow{2}{*}{\textbf{Train}} &
    \multirow{2}{*}{\textbf{Vocab}} &
    \multicolumn{3}{c|}{\textbf{Joint Control}} &
    \multicolumn{8}{c}{\textbf{Per-Property Spearman}} \\
    \cmidrule(lr){3-5}
    \cmidrule(l){6-13}
    & & \textbf{NJD} & \textbf{Tanimoto} & \textbf{Partial (/7)}
    & \textbf{logP} & \textbf{MW} & \textbf{QED} & \textbf{TPSA} & \textbf{HBD} & \textbf{HBA} & \textbf{RotBonds} & \textbf{Avg.} \\
    & & ($\downarrow$) & ($\uparrow$) & ($\uparrow$)
    & ($\uparrow$) & ($\uparrow$) & ($\uparrow$) & ($\uparrow$) & ($\uparrow$) & ($\uparrow$) & ($\uparrow$) & ($\uparrow$) \\
    \midrule
    \multirow{3}{*}{10k} & 500
      & 1.08 & 0.3165 & 6.62
      & 0.9126 & 0.7919 & 0.8381 & 0.8632 & 0.8996 & 0.8538 & 0.8032 & 0.8518 \\
      & 1000
      & 0.95 & 0.3338 & 6.75
      & 0.9115 & 0.8711 & 0.8852 & 0.9194 & 0.8839 & 0.9225 & 0.8891 & 0.8975 \\
      & 2000
      & 1.00 & 0.3206 & 6.70
      & 0.9030 & 0.8290 & 0.8287 & 0.8965 & 0.8770 & 0.9069 & 0.8870 & 0.8755 \\
    \midrule
    \multirow{3}{*}{100k} & 500
      & 0.80 & 0.3349 & 6.75
      & 0.9412 & 0.8488 & \textbf{0.9086} & \textbf{0.9469} & 0.8826 & 0.9324 & 0.8958 & 0.9080 \\
      & 1000
      & 0.81 & 0.3503 & 6.73
      & 0.9370 & 0.8687 & 0.8599 & 0.9137 & 0.8873 & 0.9273 & 0.9049 & 0.8998 \\
      & \textbf{2000}
      & \textbf{0.71} & \textbf{0.3565} & \textbf{6.82}
      & 0.9387 & \textbf{0.8966} & 0.9062 & 0.9377 & \textbf{0.9173} & \textbf{0.9458} & \textbf{0.9327} & \textbf{0.9250} \\
    \midrule
    \multirow{3}{*}{1M} & 500
      & 0.98 & 0.3373 & 6.59
      & 0.9226 & 0.8359 & 0.8515 & 0.8974 & 0.8653 & 0.9062 & 0.8639 & 0.8775 \\
      & 1000
      & 0.83 & 0.3445 & 6.76
      & 0.9291 & 0.8553 & 0.8658 & 0.9332 & 0.9089 & 0.9266 & 0.9224 & 0.9059 \\
      & 2000
      & 0.83 & 0.3313 & 6.75
      & 0.9376 & 0.8622 & 0.8924 & 0.9351 & 0.8953 & 0.9366 & 0.9212 & 0.9115 \\
    \bottomrule
  \end{tabular}
  }
  \label{tab:decoration_results}
\end{table*}

Table~\ref{tab:decoration_results} shows that decoration remains a highly controlled local-edit task. Increasing the corpus from 10k to 100k consistently improves the aggregate metrics: NJD decreases, Tanimoto similarity increases, partial satisfaction stays near or above 6.7 out of 7, and the average per-property correlation rises for every vocabulary size. The best overall setting is 100k training molecules with a 2000-fragment vocabulary, which gives the lowest NJD (0.71), highest Tanimoto similarity (0.3565), highest partial satisfaction (6.82/7), and highest average correlation (0.9250). A larger 1M corpus remains competitive but does not improve the best aggregate result, indicating that more data is not the main bottleneck once the decoration model has enough matched scaffold--branch coverage. Vocabulary size has a milder effect than corpus scale: larger vocabularies help most at 100k, but the trend is not monotonic at 10k or 1M. Across settings, high correlations for TPSA, HBA, HBD, and rotatable bonds indicate that site-conditioned retrieval can steer decorated molecules toward the requested property profile while preserving the fixed scaffold core.

\section{Detailed Fraglingo Model Design}
\label{app:model}

\begin{figure*}[!h]
    \centering
    \begin{minipage}[t]{0.48\textwidth}
        \vspace{0pt}
        The appendix expands the concise method description in Section~\ref{sec:method}. We first specify the autoregressive training examples, then detail the encoder, readout, conditioning, predictor, RVQ regularizer, losses, metrics, and model scale.

\subsection{Generation Framing}
\label{app:generation-framing}

        We decompose each molecule into fragments using BRICS bond-breaking
        rules, yielding a fragment tree with associated attachment-point
        wildcards. For each molecule, we identify the main chain as the
        longest path in this fragment tree and create two next-fragment
        prediction trajectories, one starting from each end of the chain.
        Within each trajectory, autoregressive training examples are ordered
        by a breadth-first traversal that starts from the selected terminal
        cap fragment, defined as a fragment with exactly one wildcard.
\\\\
        At step $t$, the model receives the partial assembly $G_t$ with one
        designated active wildcard atom and must predict the embedding of
        the fragment that attaches at that site. A terminal expansion is
        represented by attaching a cap fragment with a single wildcard,
        which consumes the active wildcard without introducing another
        growable site. Molecule growth stops when the assembly has no
        remaining wildcard atoms available for expansion.

    \subsection{Graph Encoder: FragmentGPS}
\label{app:encoder}

All graph-structured inputs, including both partial assemblies and individual fragments, are encoded by a single shared encoder $\phi$, FragmentGPS. FragmentGPS is a 6-layer GPS (Graph + Transformer hybrid) network with hidden dimension $d = 256$.
\\\\
Each GPSLayer contains three sub-modules applied sequentially with residual connections and LayerNorm:
    \end{minipage}
    \hfill
    \begin{minipage}[t]{0.48\textwidth}
        \vspace{0pt}
        \centering
        \includegraphics[width=\linewidth]{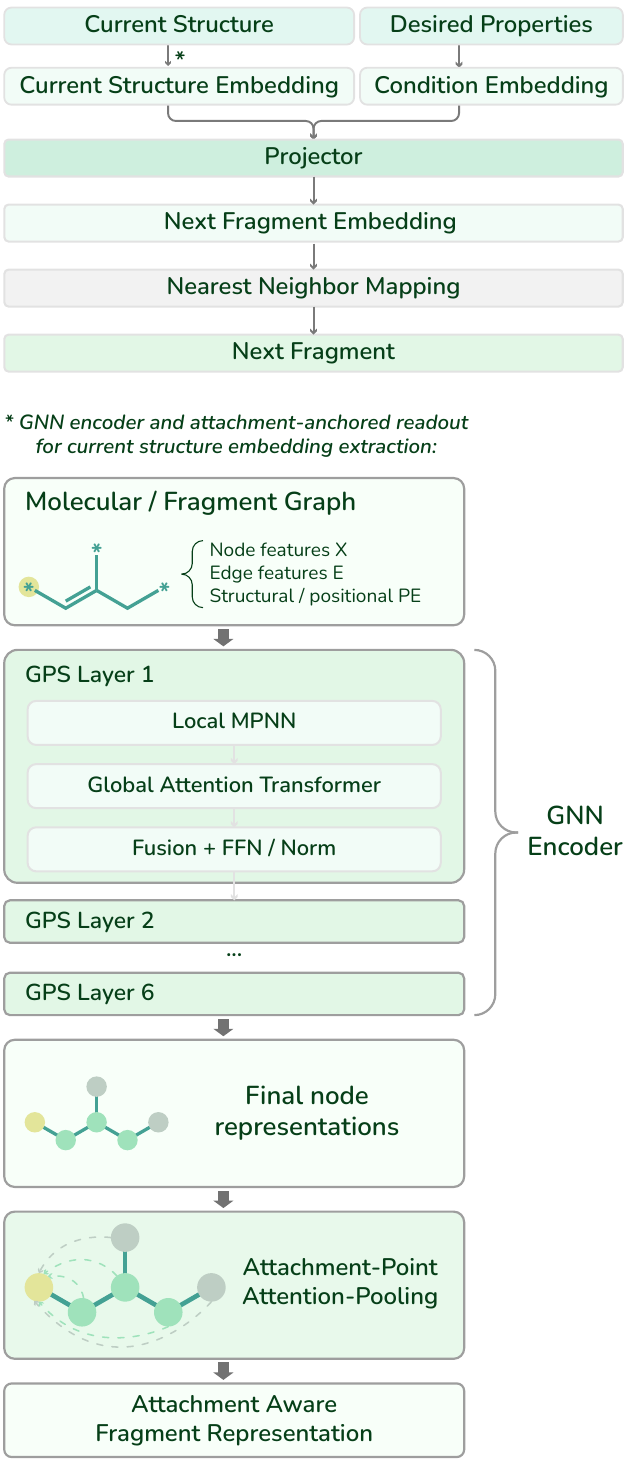}
        \caption{Detailed architecture of Fraglingo's next-fragment generation module.}
        \label{fig:model}
    \end{minipage}
\end{figure*}

\begin{enumerate}
    \item \textbf{Local message passing} via GINE (GIN with edge features): each atom aggregates messages from its bonded neighbors, where bond-type features are incorporated into the aggregation.
    \item \textbf{Global self-attention}: multi-head self-attention ($h = 8$ heads) over all atoms in the graph, giving each atom access to global context regardless of graph distance.
    \item \textbf{Feed-forward block}: a position-wise MLP ($d \to 4d \to d$, GELU activation).
\end{enumerate}
Dropout of 0.1 is applied throughout.

\paragraph{Attachment-anchored readout.}
Standard graph-level pooling (mean or sum over atoms) is replaced by an attachment-anchored readout: the graph embedding is computed as the attention-weighted sum of context-atom representations, where the query vector comes from the active attachment atom. For autoregressive generation this query atom is the open wildcard; for scaffold decoration it is the active scaffold atom. Formally, let $h_q$ denote the active query atom's final-layer hidden state and $H = [h_1, \ldots, h_n]$ the matrix of context-atom hidden states. The graph embedding is:
\begin{equation}
\label{eq:readout}
    e = \operatorname{softmax}\!\left(\frac{h_q H^\top}{\sqrt{d}}\right) H
\end{equation}
This anchors the assembly representation to the current attachment point rather than averaging over the whole molecule. For terminal cap-fragment prediction, the active wildcard atom's embedding is used directly as the query.

\subsection{Property Conditioning}
\label{app:property-conditioning}

Seven molecular property descriptors are computed per molecule: logP, molecular weight (MW), QED, topological polar surface area (TPSA), hydrogen bond donors (HBD), hydrogen bond acceptors (HBA), and the count of rotatable bonds. Each is z-scored using per-dataset statistics computed over the training set.

The property vector $\mathbf{p} \in \mathbb{R}^7$ is linearly projected to $\mathbb{R}^d$ and added to the partial-assembly embedding before the predictor MLP:
\begin{equation}
\label{eq:property-cond-app}
    e_1^{\text{cond}} = e_1 + W_p \mathbf{p} + \mathbf{b}_p
\end{equation}
This is an additive conditioning injection. During training, $\mathbf{p}$ is zeroed with probability $p_{\text{drop}} = 0.2$ independently per training example, jointly training a conditional and an unconditional model in a single pass. This enables classifier-free guidance at sampling time, where the final predicted embedding is interpolated as:
\begin{equation}
\label{eq:cfg}
    \hat{z}_{\text{guided}} = (1 + w)\,\hat{z}_{\text{cond}} - w\,\hat{z}_{\text{uncond}}
\end{equation}
for guidance scale $w \geq 0$.

\subsection{Conditioning Property Descriptors}
\label{app:property-descriptors}

We condition on seven RDKit descriptors that cover complementary aspects of molecular design:
\begin{itemize}[leftmargin=*]
    \item \textbf{logP} estimates octanol--water partitioning and serves as a proxy for hydrophobicity and membrane permeability.
    \item \textbf{MW} is molecular weight, controlling overall molecule size and often changing approximately additively with fragment additions.
    \item \textbf{QED} summarizes drug-likeness by combining several physicochemical preferences into a single score.
    \item \textbf{TPSA} measures topological polar surface area, reflecting polar atom exposure and transport-related behavior.
    \item \textbf{HBD} and \textbf{HBA} count hydrogen-bond donors and acceptors, which affect polarity and binding interactions.
    \item \textbf{RotBonds} counts rotatable bonds and acts as a compact measure of molecular flexibility.
\end{itemize}
MW, HBD, and HBA are largely determined by the fragments present, whereas logP, QED, TPSA, and RotBonds also depend strongly on how fragments are assembled. Together, these descriptors test whether the model can control both fragment-compositional and structure-dependent properties.

\subsection{\texorpdfstring{Predictor: $\text{MLP}_\text{Mapper}$}{Predictor: MLP Mapper}}
\label{app:predictor}

The conditioned context embedding $e_1^{\text{cond}}$ is passed through $\text{MLP}_\text{Mapper}$, a two-layer MLP:
\begin{equation}
\label{eq:predictor}
    \hat{z} = W_2 \cdot \operatorname{GELU}(W_1\, e_1^{\text{cond}} + b_1) + b_2
\end{equation}
with hidden dimension $4d = 1024$. The output $\hat{z} \in \mathbb{R}^d$ is the predicted embedding of the next fragment, in the same space as fragment embeddings produced by $\phi$.

\subsection{Discrete Bottleneck: RVQCodebook}
\label{app:rvq}

A 4-stage residual vector quantizer (RVQ) operates on the predicted embedding $\hat{z}$ during training. Each stage maintains a codebook of 64 vectors updated via exponential moving averages (EMA), with dead-code reset when a code goes unused for too many steps. Given $\hat{z}$, stage 1 finds the nearest codebook vector $c_1$ and passes the residual $\hat{z} - c_1$ to stage 2; after stage $m$, the residual is $\hat{z} - \sum_{\ell=1}^{m} c_\ell$. The resulting 4-tuple of code indices $(k_1, k_2, k_3, k_4)$ spans $64^4 \approx 16\text{M}$ possible combinations.

The RVQ is a training-time regularizer only and plays no role in decoding. At inference, the normalized predicted embedding $\hat{z}$ is compared directly against a precomputed table of normalized fragment embeddings $\{\phi(f)\}_{f \in \mathcal{V}}$ via inner product, which is equivalent to cosine similarity, and the fragment with the highest similarity is selected. The table can be recomputed for a larger inference vocabulary than the one used to form training targets, allowing vocabulary expansion without changing the predictor or adding a new output head.

\subsection{Training Objective}
\label{app:objective}

The model is trained end-to-end with three losses, weighted by hyperparameters:
\begin{equation}
\label{eq:app-loss}
    \mathcal{L} = \lambda_{\text{NCE}}\mathcal{L}_{\text{InfoNCE}} + \lambda_{\text{cls}}\mathcal{L}_{\text{cls}} + \lambda_{\text{commit}}\mathcal{L}_{\text{commit}}
\end{equation}
We select these weights by grid search and use $\lambda_{\text{NCE}}:\lambda_{\text{cls}}:\lambda_{\text{commit}}=1:0.5:0.25$ in the reported experiments.

\paragraph{InfoNCE loss.}
For a batch of $B$ training steps with $\ell_2$-normalized predicted embeddings $\{\hat{z}_i\}$ and $\ell_2$-normalized true next-fragment embeddings $\{e_{2,i}\}$, the InfoNCE loss is:
\begin{equation}
\label{eq:infonce}
    \mathcal{L}_{\text{InfoNCE}} = -\frac{1}{B}\sum_{i=1}^{B} \log \frac{\exp(\hat{z}_i \cdot e_{2,i} / \tau)}{\sum_{j=1}^{B} \exp(\hat{z}_i \cdot e_{2,j} / \tau)}
\end{equation}
using in-batch negatives. This is the primary signal driving the predictor to point toward the correct fragment in embedding space.

\paragraph{Fragment classification loss.}
A linear classification head is applied to the true fragment embedding $e_2 = \phi(f_{t+1})$ and trained with cross-entropy over the vocabulary. This auxiliary loss discourages embedding collapse: without it, InfoNCE alone may not keep fragment embeddings sufficiently discriminative, and the encoder can map multiple fragments to the same region of embedding space.

\paragraph{VQ commit loss.}
The commit loss pulls the continuous prediction $\hat{z}$ toward its nearest codebook vector:
\begin{equation}
\label{eq:commit}
    \mathcal{L}_{\text{commit}} = \bigl\|\hat{z} - \operatorname{sg}(\hat{z}_{\text{quantized}})\bigr\|^2
\end{equation}
where $\operatorname{sg}(\cdot)$ denotes the stop-gradient operator. Codebook vectors are updated via EMA rather than backpropagation.

\subsection{Evaluation Metrics}
\label{app:metrics}

Three metrics are tracked during training:
\begin{itemize}
    \item \textbf{acc($e_2$)}: fragment classifier accuracy on the true next-fragment embedding $e_2$, serving as a sanity check that the encoder produces discriminative representations across the vocabulary.
    \item \textbf{acc($\hat{z}$@1)}: top-1 accuracy of the predicted embedding $\hat{z}$ under the fragment classifier, measuring whether the prediction lands on the correct next fragment. This is the primary generation accuracy metric compared across ablation arms.
    \item \textbf{acc($\hat{z}$@5)}: top-5 accuracy of $\hat{z}$ under the same classifier.
\end{itemize}

\subsection{Model Scale}
\label{app:scale}

The full model has approximately 6.4M trainable parameters. The dominant components are the 6-layer FragmentGPS encoder (shared across partial assemblies and fragments), the $\text{MLP}_\text{Mapper}$ predictor head, and the RVQ codebook parameters. Table~\ref{tab:model_configs} summarizes the small, baseline, and large model configurations used to assess the effect of model scale.

\begin{table}[!hbt]
\centering
\setlength{\tabcolsep}{15pt}
\resizebox{0.6\linewidth}{!}{%
\begin{tabular}{lcccc}
\hline
\textbf{Model} & \textbf{Dim.} & \textbf{Layers} & \textbf{Heads} & \textbf{Params} \\
\hline
\textbf{Small}    & 128 & 4 & 4 & \textbf{1.2M} \\
\textbf{Baseline} & 256 & 6 & 8 & \textbf{6.4M} \\
\textbf{Large}    & 512 & 8 & 8 & \textbf{32.9M} \\
\hline
\end{tabular}%
}
\caption{Model configurations and parameter counts.}
\label{tab:model_configs}
\end{table}

Table~\ref{tab:model_size} reports property-conditional generation across model sizes and fragmentation methods. Scaling from 1.2M to 6.4M parameters yields modest improvements in joint control (NJD $2.011 \to 1.981$, Joint@$1\times$ $38.9 \to 40.88$, Joint@$2\times$ $84.9 \to 85.90$), while generation quality remains saturated. Further scaling to 32.9M does not improve joint control (NJD $2.017$, Joint@$1\times$ $39.05$, Joint@$2\times$ $83.70$, Avg. $0.733$), suggesting that this 10k-molecule, 500-fragment benchmark is not primarily capacity-limited. We also report a variant (6.4M$^\ast$) that keeps the architecture fixed but applies BPE on BRICS fragments, called BFE (BRICS-based Fragment Enumeration) in MolLingo~\citep{nguyen2026mollingo}, coarsening the fragment vocabulary into fewer, larger composite units. This variant substantially improves joint control (NJD $1.493$, Joint@$1\times$ $69.65$, Joint@$2\times$ $97.75$, Partial $6.56$) and achieves the highest average per-property Spearman correlation ($0.794$), showing that Fraglingo can benefit substantially from coarser fragmentation schemes. Thus, the plain-BRICS results reported in this work should not be interpreted as an upper bound on the retrieval framework. We keep plain BRICS fragments as the default because they are well-established and widely used; MolLingo-style BFE (BPE-on-BRICS) is an orthogonal tokenization change that composes with the same retrieval primitive, and we report it as evidence that the method can benefit from coarser vocabularies rather than as a separate model.

\begin{table*}[!hbt]
  \centering
  \caption{Effect of model size on property-conditional molecule generation (10k corpus, 500 fragments). Scaling parameters improves joint control at saturated generation quality. The 6.4M$^\ast$ row applies BPE on BRICS fragments, called BFE (BRICS-based Fragment Enumeration) in MolLingo, coarsening the vocabulary.}
  \setlength{\tabcolsep}{3pt}
  \resizebox{\linewidth}{!}{
  \begin{tabular}{l|cccc|cccc|cccccccc}
    \toprule
    \multirow{3}{*}{\textbf{Model size}} &
    \multicolumn{4}{c|}{\textbf{Generation Quality}} &
    \multicolumn{4}{c|}{\textbf{Joint Control}} &
    \multicolumn{8}{c}{\textbf{Per-Property Spearman}} \\
    \cmidrule(lr){2-5}
    \cmidrule(lr){6-9}
    \cmidrule(l){10-17}
    & \textbf{Validity}
    & \textbf{Uniqueness}
    & \textbf{Novelty}
    & \textbf{Diversity}
    & \textbf{NJD}
    & \textbf{Joint@1$\times$}
    & \textbf{Joint@2$\times$}
    & \textbf{Partial (/7)}
    & \textbf{logP}
    & \textbf{MW}
    & \textbf{QED}
    & \textbf{TPSA}
    & \textbf{HBD}
    & \textbf{HBA}
    & \textbf{RotBonds}
    & \textbf{Avg.} \\
    & ($\uparrow$)
    & ($\uparrow$)
    & ($\downarrow$)
    & ($\uparrow$)
    & ($\downarrow$)
    & ($\uparrow$)
    & ($\uparrow$)
    & ($\uparrow$)
    & ($\uparrow$)
    & ($\uparrow$)
    & ($\uparrow$)
    & ($\uparrow$)
    & ($\uparrow$)
    & ($\uparrow$)
    & ($\uparrow$)
    & ($\uparrow$) \\
    \midrule
    1.2M
      & \textbf{100\%} & \textbf{100\%} & \textbf{100\%} & \underline{0.843}
      & 2.011 & 38.9\% & 84.9\% & \underline{5.80}
      & \underline{0.823} & 0.611 & 0.568 & 0.756 & 0.821 & 0.776 & 0.756 & 0.730 \\
    6.4M
      & \textbf{100\%} & \textbf{100\%} & \textbf{100\%} & \textbf{0.8444}
      & \underline{1.9808} & \underline{40.88\%} & \underline{85.90\%} & \underline{5.80}
      & \underline{0.8373} & \underline{0.6423} & 0.5632 & \underline{0.7605} & 0.8161 & 0.7800 & \underline{0.7760} & \underline{0.7393} \\
    32.9M
      & \textbf{100\%} & \textbf{100\%} & \textbf{100\%} & \underline{0.8443}
      & 2.017 & 39.05\% & 83.70\% & 5.73
      & 0.793 & 0.621 & \underline{0.603} & 0.754 & \underline{0.825} & \underline{0.782} & 0.750 & 0.733 \\
    \midrule
    6.4M$^\ast$
      & \textbf{100\%} & \underline{99.9\%} & \textbf{100\%} & 0.8218
      & \textbf{1.4928} & \textbf{69.65\%} & \textbf{97.75\%} & \textbf{6.56}
      & \textbf{0.8710} & \textbf{0.7746} & \textbf{0.6373} & \textbf{0.8296}
      & \textbf{0.8284} & \textbf{0.8190} & \textbf{0.7993} & \textbf{0.7942} \\
    \bottomrule
  \end{tabular}
  }
  \vspace{2pt}
  \footnotesize{$^\ast$ Same architecture as 6.4M, but tokenization applies BPE on BRICS fragments, called BFE (BRICS-based Fragment Enumeration) in MolLingo, rather than using BRICS fragments alone.}
  \label{tab:model_size}
\end{table*}

\section{Baseline Experimental Setup}
\label{app:baseline-setup}

\textbf{Shared property protocol.}
All baselines use the same 7-property target vector as Fraglingo: logP, molecular weight, QED, TPSA, HBD, HBA, and rotatable bond count, with all labels computed by RDKit. Unless a baseline required a different data interface, we trained it from scratch on the same approximately 10k-molecule corpus used for the Fraglingo conditional generator, using a 90/10 train/validation split. Conditional-generation methods were evaluated on the same 100 held-out property targets, with 20 samples drawn for each target. Optimization methods were evaluated on 200 held-out source molecules paired with target 7-property vectors and scored using the same normalized joint-distance protocol as Fraglingo.

We selected baselines according to two criteria: (1) the official implementation provides a reproducible training and evaluation pipeline, and (2) the method addresses at least one of the four molecular design tasks considered in this work, namely property-conditional generation, molecule optimization, scaffold generation, or scaffold decoration. Methods whose official implementations do not support end-to-end reproduction were excluded. For example, although FragGPT~\citep{yue2024unlocking} is closely aligned with the tasks studied here, its public repository does not provide a complete training pipeline, preventing a fair comparison. We likewise excluded methods designed solely for unconditional molecular generation (such as ChemGPT~\citep{frey2023neural}, MolBART~\citep{lane2025machine}, and Chemformer~\citep{irwin2022chemformer}), as they cannot be directly adapted to our task formulations.

\paragraph{Fraglingo.} All Fraglingo variants share the same backbone architecture. The model uses a shared FragmentGPS graph encoder with a residual vector-quantized (RVQ) codebook consisting of 4 stages with 64 codes each, a commitment weight of $\beta=0.25$, and a classification loss weight of 0.5. Unless otherwise specified, architecture hyperparameters are fixed at $d_{\mathrm{model}}=256$, 6 GNN layers, 8 attention heads, and a dropout rate of 0.1. Models are trained using AdamW (learning rate $3\times10^{-4}$, weight decay $10^{-2}$) with cosine annealing, gradient clipping at 1.0, and a batch size of 32. Training runs for up to 50 epochs with early stopping (patience (S)= 3, activated after the first epoch) based on validation top-1 acc(z@1).
\subsection{Conditional generation baselines.}

\textbf{MolGPT.}
We used MolGPT~\citep{bagal2021molgpt}, a GPT-style autoregressive SMILES transformer, from the authors' official implementation\footnote{\url{https://github.com/devalab/molgpt}} without architectural changes. The model is a decoder-only causal transformer with 8 layers, 8 attention heads, and embedding dimension 256. Following the original conditioning scheme, the 7-property vector is linearly projected to a single embedding token and prepended to the SMILES sequence before the causal self-attention blocks. We trained MolGPT with AdamW, cosine learning-rate decay, and batch size 256. At inference, SMILES were sampled autoregressively with ancestral sampling at temperature 1.0 and no top-$k$ filtering until an end-of-sequence token was emitted.

\textbf{MolRWKV.}
We used MolRWKV~\citep{li2025molrwkv}, a SMILES language model based on the linear-attention RWKV-6 architecture, from the authors' official implementation\footnote{\url{https://github.com/bigwestHan/MolRWKV}} without architectural changes. The only local change was a path-resolution fix for locating custom CUDA kernel sources at build time. The model uses a 6-layer RWKV decoder with embedding dimension 256 over a 25-symbol character-level SMILES vocabulary. Conditioning is handled by a 2-layer RWKV property encoder: each scalar property is projected to the model dimension, producing 7 property-token embeddings that are encoded and injected into every decoder block. We trained MolRWKV with Adam, learning rate $5\times 10^{-4}$, batch size 128, and early stopping with patience 5 for up to 50 epochs. At inference, SMILES were sampled autoregressively until an end-of-sequence token was emitted.

\textbf{PS-VAE.}
We used PS-VAE~\citep{kong2022molecule}, a principal-subgraph variational autoencoder that assembles molecules from automatically mined fragment pieces, from the authors' official implementation\footnote{\url{https://github.com/THUNLP-MT/PS-VAE}}. Local changes were limited to PyTorch Lightning API-compatibility updates and property-scoring utilities. The released code supports QED, synthetic accessibility, and penalized logP; we added RDKit scoring and normalization ranges for molecular weight, TPSA, HBD, HBA, and rotatable bond count, and replaced penalized logP with plain RDKit logP. The encoder, piece-based VAE decoder, and property-predictor MLP were otherwise unchanged. The model used latent dimension 56 and a 500-piece vocabulary mined from the training corpus. We jointly trained the property predictor to regress all 7 properties with Adam, learning rate $10^{-3}$, batch size 32, and early stopping with patience 3 for up to 8 epochs. Because PS-VAE is not end-to-end conditional, inference used post-hoc latent optimization. For each target, we sampled 40 prior latents, optimized each for up to 50 Adam steps at learning rate 0.05 against the min-max-normalized property target with patience 10, and decoded optimized latents until 20 valid candidates were collected.

\subsection{Optimization baselines.}

\textbf{HierVAE.}
We used the HierVAE~\citep{jin2020hierarchical} hierarchical graph-to-graph translation model (HierVGNN), which encodes and decodes molecular graphs over automatically extracted structural motifs, using the authors' official implementation\footnote{\url{https://github.com/wengong-jin/hgraph2graph}}. Local changes were limited to robustness fixes: a guard against division-by-zero when preprocessing small datasets and numpy/tensor-to-float coercions for logging. We trained the translator variant with hidden and embedding size 270, latent size 4, and graph/tree message-passing depth 20. Training used the same matched-molecule-pair corpus as Fraglingo optimization: 10k source-target pairs from the full matched-pairs dataset. Molecules incompatible with the codebase's motif extraction were filtered out. We trained with Adam, learning rate $10^{-3}$, batch size 32, 20 epochs, and KL weight 0.3 annealed at rate 0.9. At inference, each held-out source used 20 stochastic translations over multiple motif attachment roots.

\end{document}